\documentclass{article}

\usepackage[main, final]{neurips_2026}

\usepackage[utf8]{inputenc} 
\usepackage[T1]{fontenc}    
\usepackage{hyperref}       
\usepackage{url}            
\usepackage{booktabs}       
\usepackage{amsfonts}       
\usepackage{nicefrac}       
\usepackage{microtype}      
\usepackage{xcolor}         
\usepackage{amsmath}
\usepackage{graphicx}
\usepackage{booktabs}
\usepackage{algorithm}
\usepackage{algpseudocode}
\usepackage{makecell}
\title{Interactive 4D Volumetric Liquid Forecasting under Moving-Solid Interaction}

\author{%
  Jongwon Ryu , \quad
  Sunjae Yoon , \quad
  Junyeong Kim\thanks{Corresponding author.} \\
  Department of Artificial Intelligence \\
  Chung-Ang University \\
  \texttt{\{fbwhddnjs511,sunjaeyoon,junyeongkim\}@cau.ac.kr}
}

\begin{document}

\maketitle



\begin{abstract}
We propose the task of \emph{Interactive 4D Volumetric Liquid Forecasting}: predicting the spatiotemporal evolution of a future dense 3D liquid velocity field conditioned on how a solid object moves through it.
This task provides an upstream physical forecasting primitive for systems that need to reason about motion-conditioned liquid response, including robotic liquid manipulation, digital twins, and rigid-fluid video generation.
Existing works have made efforts in predicting flow around static or prescribed solid geometries and synthesizing visually plausible fluid motion. Yet these approaches provide limited coverage for interactive liquid dynamics, as they are not capable of forecasting how a moving solid induces a future full-domain liquid response.
In this forecasting scenario, the hydrodynamic effect is condensed near an evolving liquid-solid interface, but the prediction target is a global spatiotemporal rollout.
To address this challenge, we propose a two-level recurrent forecasting framework, \textbf{TIDE}/\textbf{TIDES}: TIDE isolates localized interface-driven forecasting, while TIDES adds transport-aware coupling and explicit stabilization to reduce accumulated rollout error.
Across adapted neighboring baselines, controlled component ablations, motion/size OOD tests, and long-horizon evaluation, TIDES achieves the best forecasting performance and the ablations identify localized interaction, transport coupling, and stabilization as the key contributors.
Together, our task formulation, benchmark, and TIDE/TIDES study establish a controlled foundation for moving-solid-conditioned dense liquid forecasting. 
The code is publicly available at \url{https://github.com/jongwonryu/TIDE-TIDES}.

\end{abstract}


\section{Introduction}

When a solid object moves through liquid, its motion first creates disturbances near the evolving solid--liquid interface, and these local disturbances propagate into a global flow response over time~\citep{kolahdouz2021sharp,xiao2024fourier,tao2026neural,zhong2025benchmarking}. 
We formalize this setting as the task of \emph{interactive 4D volumetric liquid forecasting}: given an initial liquid state and a future moving-solid trajectory, the goal is to predict the future dense 3D liquid velocity field over the full domain. 
This task asks whether a model can infer not only where liquid particles move, but also how the full volumetric liquid state evolves as an explicit physical field when a solid object moves, pushes, stirs, or passes through the liquid.

This capability provides an upstream physical forecasting primitive for interactive systems that must reason about motion-conditioned liquid response. 
In robotic liquid manipulation, a system may need to compare candidate pouring, stirring, scooping, or pushing motions before acting~\citep{midlagajni2025learning}. 
In digital twins and simulation-driven design, operators may wish to ask how changes in a moving tool, actuator, or trajectory alter the surrounding flow state~\citep{kochkov2021machine,li2023geometry}. 
In rigid--fluid video generation and future 4D world models, realistic interaction requires not only moving the rigid object correctly, but also producing coherent wakes, waves, and other liquid responses induced by that motion~\citep{akkerman2025interdyn,shi2024motion,jia2025learning}. 
These settings motivate a reusable forecasting primitive for repeated querying: many candidate solid motions can be evaluated under a shared physical scene.

Existing learned flow models address related but incomplete settings. 
Neural operators and CFD surrogates have made substantial progress in efficient field prediction, operator learning, and geometry-aware surrogate modeling~\citep{li2020fourier,kovachki2023neural,rahman2022u,wen2022u,tran2021factorized,li2023fourier}. 
However, these methods typically focus on operator regression, geometry-conditioned prediction, or flow around static or prescribed solid geometries, rather than forecasting the evolving liquid response induced by a moving solid. 
Volumetric flow enhancement methods learn dense velocity fields, but usually reconstruct or enhance already observed measurements rather than predict future motion-conditioned liquid response~\citep{ferdian20204dflownet,rutkowski2021enhancement,suk2026physics}. 
Visual rigid--fluid synthesis methods can generate plausible fluid appearance, but they often do not provide an explicit dense physical state that can be reused for rollout, querying, or downstream reasoning~\citep{akkerman2025interdyn,shi2024motion,jia2025learning,du2024conditional}. 
Particle-, mesh-, and graph-based learned dynamics capture local interactions in simulator-native representations, but their primary prediction space is not a full-domain dense Eulerian volume~\citep{pfaff2020learning,gaoamesh}. 
Similarly, scene-flow methods estimate 3D motion fields, but they target scene or point motion rather than physically explicit liquid-state forecasting~\citep{li2023fast,vedder2024neural,liang2025zero}. 
Together, these limitations open the problem of forecasting full-domain liquid response under time-varying solid interaction.

The key challenge is local-to-global forecasting: the solid directly affects the liquid near the moving solid--liquid interface, but this local disturbance propagates through the full domain and must be predicted as a dense Eulerian velocity rollout.
The model must therefore forecast the future dense Eulerian velocity field, while remaining stable under autoregressive rollout and distribution shifts in motion, shape, and horizon~\citep{koehler2024apebench,zhong2025benchmarking,shaffer2026structure}. 

To address these challenges, we introduce a benchmark for moving-solid-conditioned dense liquid forecasting and propose a recurrent forecasting framework with two levels. 
\textbf{TIDE} (\emph{T}emporal \emph{I}nterface-\emph{D}riven liquid field \emph{E}stimator) isolates the localized interface-driven forecasting principle: solid-induced updates should be generated where the interaction occurs and then propagated through rollout. 
\textbf{TIDES} retains this localized core and augments it with transport-aware coupling and explicit stabilization for robust long-horizon forecasting. 

Our benchmark and shared evaluation protocol are designed for this task. 
Since no standard benchmark or exact task-matched baseline exists for moving-solid-conditioned dense liquid forecasting, we compare against task-necessity baselines, adapted neighboring baselines, and controlled within-family ablations under a shared forecasting protocol, following the broader emphasis on rollout, physical fidelity, OOD generalization, and benchmarked scientific forecasting~\citep{yang2025multiphysics,hu2026realpdebench}. 

Our experiments reveal a clear modeling hierarchy. 
Localized interface-driven interaction provides the forecasting core, while transport-aware coupling and explicit stabilization improve long-horizon rollout behavior. 
Under the shared benchmark, \textbf{TIDES} achieves consistent gains in aggregate accuracy, transport fidelity, and robustness across controlled ablations, adapted neighboring baselines, motion/size OOD tests, and long-horizon evaluation shifts.

\begin{figure}[t]
    \centering
    \includegraphics[width=\linewidth]{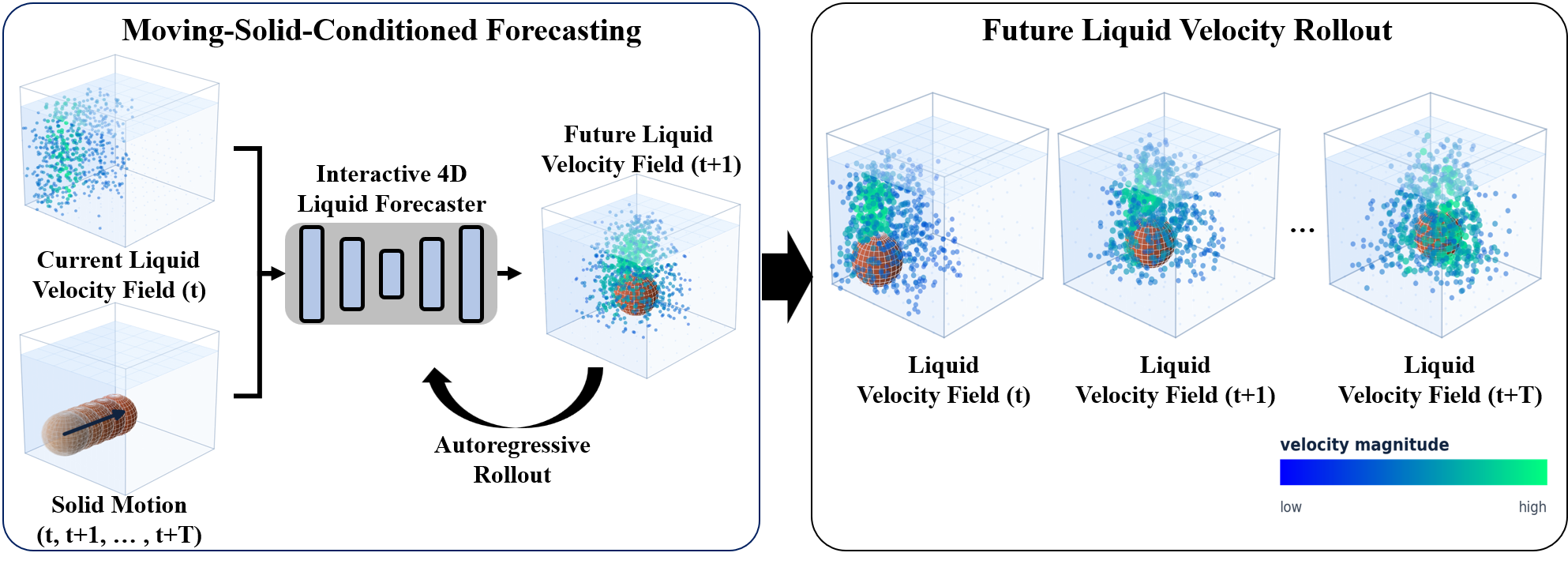}
\caption{
Task overview.
Left: moving-solid-conditioned forecasting setup, where the current liquid velocity field and the known solid motion sequence are provided to an autoregressive liquid forecaster.
Right: the corresponding future liquid velocity rollout.
}

    \label{fig:task_overview}
\end{figure}

\section{Related Work}

\paragraph{Neural fluid surrogates and neural operators.}
A large body of work studies learned surrogates for PDEs and fluid dynamics, especially through neural operator frameworks such as FNO, U-NO, U-FNO, and Geo-FNO~\citep{li2020fourier,kovachki2023neural,rahman2022u,wen2022u,li2023fourier}. These models have shown strong performance in operator regression, multiscale prediction, and geometry-aware field modeling~\citep{li2023geometry,wu2024transolver,wen2025geometry,rabeh20253d}. However, they are typically designed for operator regression or geometry-conditioned surrogate prediction. In contrast, our setting requires forecasting a dense liquid rollout under time-varying solid motion, where object--fluid exchange occurs through an evolving interaction interface.

\paragraph{Volumetric flow prediction and enhancement.}
Another related line of work learns 3D spatiotemporal velocity fields in scientific and medical settings. For example, 4DFlowNet and related methods use CFD-assisted supervision to super-resolve or denoise 4D flow MRI~\citep{ferdian20204dflownet,rutkowski2021enhancement}. These methods show that volumetric velocity fields and local structures such as near-wall behavior can be learned effectively~\citep{rutkowski2021enhancement,suk2026physics}. However, they enhance or reconstruct already observed flow measurements, whereas we forecast future liquid response conditioned on changing solid motion and occupancy.

\paragraph{Interface-aware fluid--solid learning and particle/mesh dynamics.}
Work on fluid--solid interaction highlights the importance of coupling through boundaries or interfaces rather than uniform global updates~\citep{kolahdouz2021sharp,zhong2025benchmarking,xiao2024fourier}. Classical FSI formulations and recent interface-aware learning methods both support this view~\citep{kolahdouz2021sharp,tao2026neural}. In parallel, particle-graph simulators such as GNS and mesh-based learned simulators such as MeshGraphNets provide strong inductive biases for local interaction and transport-consistent dynamics~\citep{pfaff2020learning,midlagajni2025learning}. Our work draws on these insights but targets a different prediction space: dense Eulerian liquid velocity rollout, using Lagrangian transport only as a coupled consistency mechanism rather than as the primary output.

\paragraph{Benchmarking scientific forecasting models.}
Recent benchmarks emphasize evaluating scientific forecasting beyond one-step regression accuracy~\citep{koehler2024apebench, hu2026realpdebench}. APEBench highlights rollout-based and long-horizon evaluation for PDE emulators~\citep{koehler2024apebench}, while ML4CFD stresses accuracy, physical fidelity, out-of-distribution generalization, and efficiency~\citep{yagoubi2025neurips}. This is especially relevant for interactive solid--fluid forecasting, where local interface errors can accumulate into transport or stability failures over autoregressive rollout. Since no exact task-matched benchmark exists for our setting, we evaluate adapted neighboring baselines and controlled within-family ablations under a shared protocol to isolate necessary modeling ingredients.


\begin{figure}[t]
    \centering
    \includegraphics[width=\linewidth]{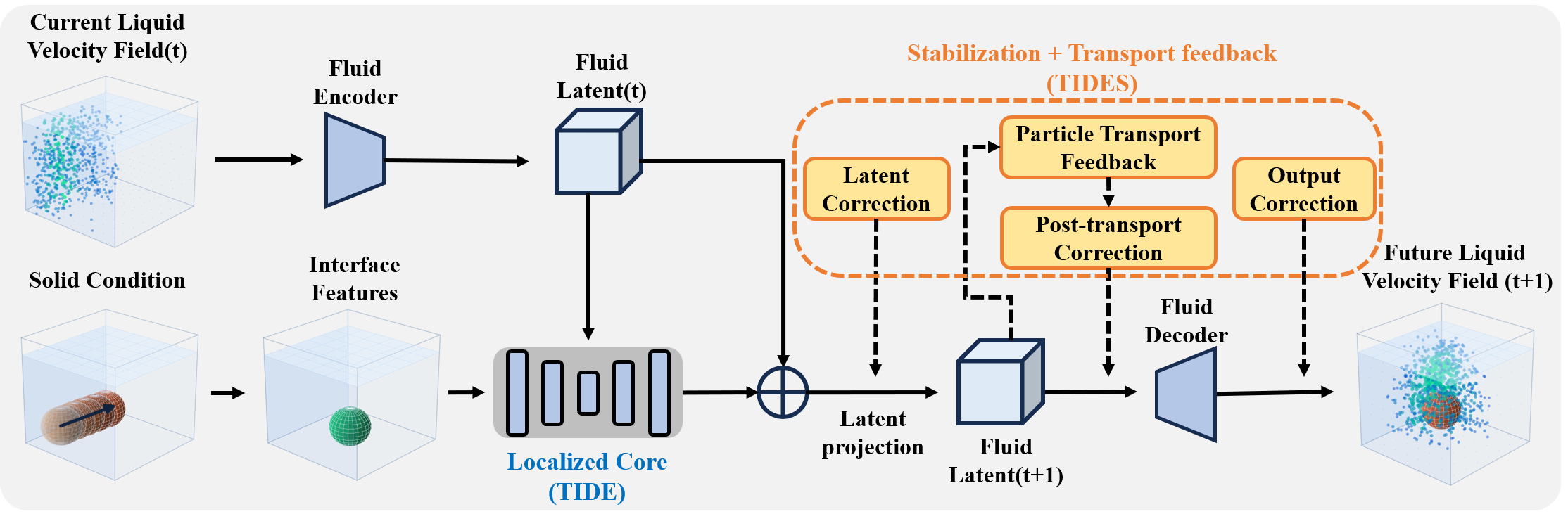}
    \caption{Method overview. TIDE provides the localized interface-driven core, while TIDES augments it with transport feedback and stabilization modules for autoregressive rollout.
}
    \label{fig3:method}
\end{figure}

\section{Method}

\subsection{Problem Setting}

We study \emph{moving-solid-conditioned interactive 4D volumetric liquid forecasting}. Given the current Eulerian liquid velocity field $u_t$ and a known future solid-state sequence, the model predicts a dense Eulerian liquid-velocity rollout over horizon $T$:
\[
\hat{u}_{t+1:t+T}
=
f_\theta(u_t, S_{t+1:t+T}),
\qquad
S_\tau=(\phi_\tau^s,v_\tau^s).
\]
Here $u_t\in\mathbb{R}^{3\times D\times H\times W}$ is the current 3D liquid velocity field, $\phi_\tau^s$ is the signed-distance field (SDF) of the moving solid, and $v_\tau^s$ is the solid velocity field rendered on the same voxel grid. SDF-based geometric representations are widely used in geometry-aware neural operators and 3D flow surrogates because they provide a continuous encoding of object boundaries and near-surface structure~\citep{rabeh20253d,li2023geometry}. The future solid sequence is treated as known exogenous conditioning; the model predicts only the liquid response. This matches repeated-query settings where candidate solid motions are provided and the goal is to forecast their induced liquid dynamics, which is closely related to rigid-body FSI settings where solid motion acts as a control signal for the surrounding fluid~\citep{zhong2025benchmarking}.

Rather than treating many geometry cues as independent inputs, we derive them deterministically from the future solid state. Specifically, from $(\phi_\tau^s,v_\tau^s)$ we compute solid occupancy, interface masks, surface normals, and boundary-aligned velocity cues:
\[
o_\tau^s=\mathbf{1}[\phi_\tau^s\le 0],
\qquad
m_\tau^\partial=\mathrm{Edge}(o_\tau^s),
\qquad
n_\tau=\nabla\phi_\tau^s.
\]
We collect these solid-derived fields into an interface representation
\[
\mathcal{I}_\tau
=
\mathcal{D}(\phi_\tau^s,v_\tau^s)
=
\{o_\tau^s,m_\tau^\partial,n_\tau,v_\tau^s\}.
\]
In implementation, these fields are voxel-aligned with the liquid grid and are encoded by lightweight 3D convolutional encoders before being used by the interaction modules. The central difficulty is local-to-global: solid-induced changes originate near the evolving solid--liquid interface, but the target is a stable full-domain Eulerian rollout. This interface-mediated coupling is a central challenge in classical and learned fluid--structure interaction~\citep{kolahdouz2021sharp,tao2026neural}.

\subsection{TIDE: Localized Interface-Driven Core}

\textbf{TIDE} isolates the core modeling principle of the task: moving-solid effects should be generated near the solid--liquid interface and then propagated through autoregressive rollout, rather than injected everywhere as unstructured global conditioning. This design follows the observation from FSI formulations that object--fluid exchange is mediated through boundary or interface conditions~\citep{kolahdouz2021sharp,zhong2025benchmarking}. We express this one-step localized forecaster as
\begin{equation}
\hat{u}_{t+1}^{\mathrm{core}}
=
\Pi\!\left(
B_\theta(u_t)
+
\Delta u_\theta^{\mathrm{int}}(u_t,\mathcal{I}_{t+1})
\right),
\label{eq:core_localized}
\end{equation}
where $B_\theta$ predicts the bulk volumetric liquid evolution, $\Delta u_\theta^{\mathrm{int}}$ predicts a solid-induced residual update conditioned on the next-step interface representation, and $\Pi$ is a lightweight projection or stabilization operator.
The two terms in Eq.~\eqref{eq:core_localized} play different roles. The bulk path models liquid evolution over the full volume, while the interface path focuses on changes caused by the moving solid. In implementation, the bulk predictor is realized with a compact 3D volumetric backbone, and the interface residual is produced from voxelized interface features using local 3D convolutional processing. This makes TIDE a controlled model for testing whether explicitly localized interface conditioning is necessary, while avoiding additional transport or correction modules used in the full model.

\subsection{TIDES: Transport-Aware Stabilized Extension}

\textbf{TIDES} retains the localized interface-driven core of TIDE, but makes it more robust for autoregressive rollout by adding transport-aware feedback and explicit stabilization. Long-horizon rollout stability is a recurring concern in learned PDE and fluid emulators, where one-step accuracy alone is often insufficient for reliable forecasting~\citep{koehler2024apebench}. It operates in latent space. Let $z_f^t$ denote the current fluid latent grid, and let $\mathcal{I}_{t+1}$ denote the derived interface representation from the next solid state.

TIDES first computes a tentative next fluid latent by combining a bulk fluid update with an interface-mediated residual:
\begin{equation}
\tilde{z}_f^{t+1}
=
z_f^t
+
\Delta z_f^{\mathrm{bulk}}
+
\Delta z_f^{\mathrm{int}}.
\end{equation}
The bulk update is produced by 3D U-NO-style backbone, $\Delta z_f^{\mathrm{bulk}}=F_\theta(z_f^t)$, which models domain-wide liquid evolution. U-shaped neural operators and U-FNO-style models have been used to improve multiscale field prediction and memory efficiency in neural operator learning~\citep{rahman2022u,wen2022u}. The interface update is produced by encoding $\mathcal{I}_{t+1}$ into an interface latent, updating the recurrent interface state, and injecting the resulting interface information back into the fluid grid through a localized mediator. This gives $\Delta z_f^{\mathrm{int}}$, the latent residual induced by the moving solid.

The tentative latent is then stabilized before transport:
\begin{equation}
\bar{z}_f^{t+1}
=
\Pi_{\mathrm{lat},1}(\tilde{z}_f^{t+1}),
\end{equation}
where $\Pi_{\mathrm{lat},1}$ is a learned residual correction operator. From this corrected latent, TIDES decodes a provisional velocity field and uses it to advect auxiliary particles. These particles are not the prediction target; they act as Lagrangian probes of the transport implied by the Eulerian latent state. Similar Lagrangian or particle-based representations have been useful for learning local transport and liquid--object interactions~\citep{pfaff2020learning,midlagajni2025learning}.

The resulting particle displacement is rasterized back to the voxel grid and fused into the fluid latent:
\begin{equation}
z_f^{t+1}
=
\Pi_{\mathrm{lat},2}
\left(
\bar{z}_f^{t+1}
+
R_{\mathrm{P2G}}(p_t,p_{t+1},\mathcal{I}_{t+1})
\right).
\end{equation}
Here, $R_{\mathrm{P2G}}$ converts particle displacement and local particle motion statistics into a grid-aligned transport feedback signal, and $\Pi_{\mathrm{lat},2}$ is a second learned correction operator applied after this feedback. This post-transport correction helps reduce error accumulation during long-horizon rollout, which is especially important in autoregressive scientific forecasting~\citep{koehler2024apebench}.

Finally, TIDES decodes the stabilized latent and applies an interface-aware output correction:
\begin{equation}
\hat{u}_{t+1}
=
\Pi_{\mathrm{out}}
\left(
D_v(z_f^{t+1}),\mathcal{I}_{t+1}
\right).
\end{equation}
The output correction is a convolutional velocity correction conditioned on the same interface representation. Thus, TIDES does not replace the localized principle of TIDE; it strengthens it with latent correction, particle-grid transport feedback, and output-level stabilization.

\subsection{Autoregressive Rollout and Objective}

The model is autoregressive only in the liquid state; the future solid sequence is supplied at every step:
\begin{equation}
\hat{u}_{t+k+1}
=
f_\theta(\hat{u}_{t+k},\phi_{t+k+1}^s,v_{t+k+1}^s,p_{t+k}),
\qquad
k=0,\dots,T-1,
\qquad
\hat{u}_{t}=u_t.
\end{equation}
Thus, the liquid state is recursively updated from the model's own predictions, while the solid trajectory remains an external conditioning signal. This follows the broader autoregressive emulator setting used in learned PDE forecasting benchmarks~\citep{koehler2024apebench}. When the dual-space branch is enabled, the particle state is updated jointly with the Eulerian rollout; otherwise, the target remains only the dense Eulerian velocity sequence.

We train with five functional loss groups:
\begin{equation}
\mathcal{L}
=
\mathcal{L}_{\mathrm{field}}
+\lambda_{\mathrm{roll}}\mathcal{L}_{\mathrm{roll}}
+\lambda_{\mathrm{int}}\mathcal{L}_{\mathrm{int}}
+\lambda_{\mathrm{trans}}\mathcal{L}_{\mathrm{trans}}
+\lambda_{\mathrm{stab}}\mathcal{L}_{\mathrm{stab}}.
\label{eq:total_loss}
\end{equation}
$\mathcal{L}_{\mathrm{field}}$ anchors one-step dense velocity prediction, $\mathcal{L}_{\mathrm{roll}}$ supervises multi-step autoregressive quality, $\mathcal{L}_{\mathrm{int}}$ emphasizes interface-local fidelity, $\mathcal{L}_{\mathrm{trans}}$ aligns Eulerian prediction with Lagrangian transport behavior, and $\mathcal{L}_{\mathrm{stab}}$ regularizes stable rollout. Rollout-based objectives and long-horizon metrics are important for evaluating temporal generalization in neural PDE emulators. In the full implementation, these groups decompose into several auxiliary losses, including dense field losses, rollout losses, interface-weighted losses, particle/transport consistency terms, and correction regularizers. Appendix~\ref{app:loss-details} gives the exact decomposition and weights.



\section{Experiments}

\subsection{Task Setup and Evaluation}
We evaluate \emph{moving-solid-conditioned interactive 4D volumetric liquid forecasting}: given an initial liquid state and a solid condition, the model predicts a dense 3D Eulerian liquid-velocity rollout on a $32 \times 128 \times 128$ grid at 8 FPS. The benchmark contains sphere, cube, and cone object families with motion factors spanning path length, speed, acceleration/deceleration, and straight or curved trajectories. We use 12800 / 1600 / 1600 / 4000 episodes for train / validation / test / $test_{\mathrm{ood}}$.

All core comparisons use the same rollout-oriented training setup, optimization budget, and evaluation protocol; the main comparison reports mean performance over 5 random seeds with 10000 update steps per seed. Models are trained and tested at rollout horizon 8, with horizon extension evaluated at rollout lengths 16 and 24. Motion OOD uses unseen combinations of motion and trajectory factors, while size OOD doubles sphere radius, cone base radius, and cube edge length at test time.

The dense target is a benchmark-defined Eulerian velocity field constructed from particle simulations by trilinear/CIC particle-to-grid conversion, not a solver-native continuum field. Exact metric definitions, fallback mask rules, and particle-to-grid details are given in Appendix~\ref{app:metrics_p2g}. Because the task is motivated by repeated querying, Appendix~\ref{app:efficiency_latency} also reports latency: \textbf{TIDES} takes 302.9\,ms for an 8-step rollout on the benchmark hardware, compared with about 6.46\,s of DualSPHysics simulation time per physical second for representative benchmark cases. We use this as a forecasting-speed reference only, not as a claim of simulator equivalence. Development-time variants such as interface-near-aware and $t{+}1$-focused objectives are excluded from the primary evidence because they change the final method definition.

Because no standard benchmark exists for this exact task, we use a metric suite adapted from autoregressive PDE forecasting, transport-sensitive evaluation, and physics-aware diagnostics. We treat rollout field error ($l_{\mathrm{roll}}$), near-interface field error ($\mathrm{near}$), and transport endpoint error ($traj_{\mathrm{end}}$) as the primary forecasting metrics, since they measure overall rollout quality, interaction-region accuracy, and material transport consistency. We additionally report far-field error ($\mathrm{far}$) for propagation behavior and divergence delta ($\Delta_{\mathrm{div}}$) as a stabilization-oriented diagnostic, not as a standalone measure of physical fidelity. Since field metrics are normalized voxel averages, we emphasize consistent ranking, cross-metric tradeoffs, and robustness under OOD and long-horizon evaluation rather than raw decimal scale alone.

\subsection{Interface-Localized Response Characterization}
\label{sec:task_characterization}

We first characterize how moving-solid interaction affects volumetric liquid response. Here, response energy denotes the per-voxel squared temporal velocity change, $\|u_t(x)-u_{t-1}(x)\|_2^2$, averaged over valid liquid voxels within each relative distance bin from the moving-solid interface. Table~\ref{tab:task_distance} shows that this response energy is concentrated near the evolving interaction interface and drops by about $455\times$ from the nearest interface band to the far field. This motivates treating the task as interface-driven forecasting rather than bulk-only prediction.
\begin{table}[t]
\centering
\caption{Distance-to-interface response energy. Lower interface distance corresponds to voxels closer to the evolving interaction boundary.}
\label{tab:task_distance}

\begin{tabular}{lc}
\toprule
Relative interface distance bin & Mean response energy \\
\midrule
$[0.00, 0.05)$ & 1.663763 \\
$[0.05, 0.10)$ & 0.344675 \\
$[0.10, 0.20)$ & 0.104469 \\
$[0.20, 0.35)$ & 0.012967 \\
$[0.35, 0.50)$ & 0.004842 \\
$[0.50, 1.00)$ & 0.003659 \\
\bottomrule
\end{tabular}
\end{table}

\subsection{Main Comparison}\label{sec:main_comparison}

Since no standard benchmark or exact task-matched baseline exists, Table~\ref{tab:main_comparison} is organized to answer complementary comparison questions rather than to present a single exact-competitor ranking. \textbf{Bulk Only} tests whether dense liquid rollout can be predicted without explicit moving-solid conditioning. The adapted \textbf{3D ResUNet}, \textbf{U-NO}, \textbf{Transolver}, and \textbf{GNS} baselines anchor the benchmark against neighboring volumetric, neural-operator, transformer-based, and particle model families~\citep{rahman2022u,wu2024transolver,pfaff2020learning,midlagajni2025learning}. Our variants then test the modeling hierarchy within the proposed family: \textbf{TIDE} uses the localized interface-driven core, \textbf{Finite Support} widens the localized residual support as an exploratory variant, and \textbf{TIDES} adds transport-aware feedback and stabilization.

All voxel models are trained under the same dense target, rollout horizon, optimization budget, and metric protocol, with \textbf{3D ResUNet} receiving a small proxy width sweep before full-budget retraining. \textbf{GNS} preserves native particle-state prediction and is evaluated by rasterizing predicted particle velocities to the common voxel grid, following the broader use of particle/graph simulators for learned dynamics~\citep{pfaff2020learning,midlagajni2025learning}. We deliberately do not attach our localized mediator, transport feedback, or correction modules to the external baselines, since doing so would turn them into backbone-swapped versions of our method rather than neighboring references. Further adaptation details are given in Appendix~\ref{app:baseline_adapt} and Table~\ref{tab:baseline_adaptation_details}.

\begin{table}[t]
\centering
\caption{Main comparison under the same optimization budget and evaluation protocol, aggregated over 5 random seeds. \textbf{Bold} indicates the best performance.}
\label{tab:main_comparison}
\begin{tabular}{lccccc}
\toprule
Model & $l_{\mathrm{roll}}\downarrow$ & $traj_{\mathrm{end}}\downarrow$ & $\mathrm{near}\downarrow$ & $\mathrm{far}\downarrow$ & $\Delta_{\mathrm{div}}$ \\
\midrule

3D ResUNet  & 0.007585 & 0.122781 & 0.005757 & 0.009553 & +0.000000 \\
U-NO  & 0.006542 & 0.112111 & 0.005807 & 0.005999 & -0.000376 \\
GNS  & 3.767170 & 5.555210 & 3.673578 & 3.957512 & +0.000000 \\
Transolver & 0.006981 & 0.142872 & 0.005712 & 0.008737 & +0.000000 \\
\midrule
Bulk Only & 0.007366 & 0.121042 & 0.005771 & 0.007495 & +0.000241 \\

Finite Support & 0.006240 & 0.113406 & 0.005855 & 0.006389 & -0.000150 \\
TIDE & 0.006113 & 0.110904 & 0.005660 & 0.006279 & -0.000248 \\
TIDES & \textbf{0.005746} & \textbf{0.104397} & \textbf{0.005581} & \textbf{0.005826} & \textbf{-0.057067} \\
\bottomrule
\end{tabular}
\end{table}

\begin{figure}[t]
    \centering
    \includegraphics[width=0.8\linewidth]{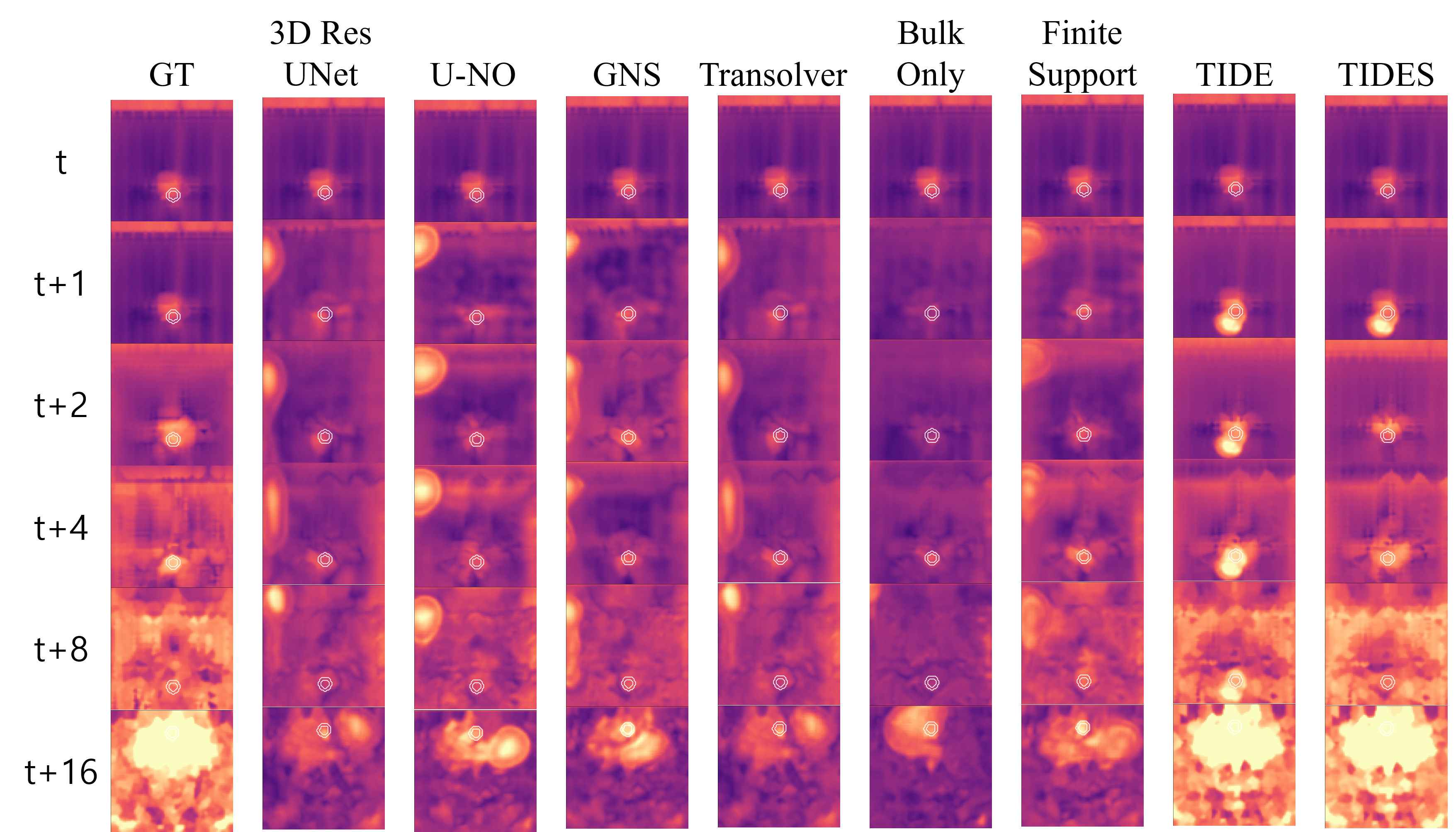}
    \caption{Qualitative comparison of autoregressive velocity rollouts.
Each column shows a model and each row shows a rollout step from $t$ to $t{+}16$.
The white contour indicates the moving-solid location,and brighter colors indicating higher velocity magnitude.
}
    \label{fig:qualitative}
\end{figure}

The main pattern is the progression from no explicit interaction modeling to localized interaction modeling and then to stabilized transport-aware rollout. \textbf{Bulk Only} is weaker than the interaction-aware variants, showing that the moving solid cannot be treated as incidental context. \textbf{TIDE} improves the primary forecasting metrics relative to the adapted external references and the widened-support variant, which supports localized interface conditioning as the core inductive bias. \textbf{TIDES} further reduces rollout, interface, far-field, and trajectory errors while producing the clearest stabilization diagnostic, suggesting that transport-aware feedback and correction modules matter once the localized core is used autoregressively.

\subsection{Localized Principle Validation}
\label{sec:localized_ablation}

We next isolate the localized-interaction design. Table~\ref{tab:localized_ablation} compares \textbf{TIDE} with variants that remove projection, remove object-local context, replace localized residual updates with global conditioning, or widen the interface support.

\begin{table}[t]
\centering
\caption{Localized-centered principle ablation. \textbf{Bold} indicates the best performance.}
\label{tab:localized_ablation}
\begin{tabular}{lccccc}
\toprule
Variant & $l_{\mathrm{roll}}\downarrow$ & $traj_{\mathrm{end}}\downarrow$ & $\mathrm{near}\downarrow$ & $\mathrm{far}\downarrow$ & $\Delta_{\mathrm{div}}$ \\
\midrule
Localized(TIDE) & \textbf{0.006113} & \textbf{0.110904} & \textbf{0.005660} & \textbf{0.006279} & \textbf{-0.000248} \\
w/o projection & 0.006943 & 0.118163 & 0.005844 & 0.006985 & +0.000000 \\
w/o objpatch & 0.006357 & 0.111277 & 0.005727 & 0.006406 & +0.000140 \\
with Global Conditioning & 0.006542 & 0.112111 & 0.006807 & 0.006599 & -0.000076 \\
Finite Support & 0.006240 & 0.113406 & 0.005855 & 0.006389 & -0.000050 \\
\bottomrule
\end{tabular}
\end{table}

\textbf{TIDE} gives the best primary metrics in this group. Removing projection produces the largest degradation, indicating that post-update consistency matters even for the localized core. Removing object-local context, replacing localized residuals with global conditioning, or widening the support also degrades performance, supporting the default narrow-band localized design.

\subsection{Full-Model Tradeoffs}
\label{sec:full_ablation}

\begin{table}[t]
\centering
\caption{Full-model component ablation. \textbf{Bold} indicates the best performance.}
\label{tab:full_ablation}
\begin{tabular}{lccccc}
\toprule
Variant & $l_{\mathrm{roll}}\downarrow$ & $traj_{\mathrm{end}}\downarrow$ & $\mathrm{near}\downarrow$ & $\mathrm{far}\downarrow$ & $\Delta_{\mathrm{div}}$ \\
\midrule
TIDES & \textbf{0.005746} & \textbf{0.104397} & \textbf{0.005581} & \textbf{0.005826} & \textbf{-0.057067} \\
w/o dual feedback & 0.005906 & 0.153737 & 0.005671 & 0.005973 & -0.009762 \\
w/o latent correction & 0.005750 & 0.109175 & 0.005642 & 0.005975 & -0.008878 \\
w/o output correction & 0.006017 & 0.108425 & 0.005662 & 0.006139 & -0.002105 \\
w/o both corrections & 0.005825 & 0.162747 & 0.005690 & 0.005837 & -0.009326 \\
\bottomrule
\end{tabular}
\end{table}

We next remove one full-model component at a time from \textbf{TIDES}: dual transport feedback, latent correction, output correction, or both correction stages together.
Show in Table \ref{tab:full_ablation}, \textbf{TIDES} is best across the primary metrics, and each removal weakens performance. The largest drop comes from removing dual feedback, which sharply worsens transport endpoint accuracy. Removing latent or output correction also degrades the metrics, while removing both corrections causes a larger transport failure. This pattern supports the view that transport feedback and correction stages are complementary rather than a single interchangeable addition.

\subsection{Objective Complementarity}
\label{sec:loss_ablation}

\begin{table}[t]
\centering
\caption{Single-loss ablation of the final Full objective on the Full Stabilized model. Each row removes one loss term while keeping the architecture fixed. \textbf{Bold} indicates the best performance.}
\label{tab:loss_ablation}
\begin{tabular}{lccccc}
\toprule
Variant & $l_{\mathrm{roll}}\downarrow$ & $traj_{\mathrm{end}}\downarrow$ & $\mathrm{near}\downarrow$ & $\mathrm{far}\downarrow$ & $\Delta_{\mathrm{div}}$ \\
\midrule
Full Loss & \textbf{0.005746} & \textbf{0.104397} & \textbf{0.005581} & \textbf{0.005826} & \textbf{-0.057067} \\
w/o $L_{\mathrm{traj}}$  & 0.005801 & 0.110317 & 0.005768 & 0.005951 & -0.030720 \\
w/o $L_{\mathrm{int}}$   & 0.005867 & 0.105963 & 0.005638 & 0.005903 & -0.045270 \\
w/o $L_{\mathrm{roll}}$  & 0.006907 & 0.105727 & 0.006012 & 0.007216 & -0.041455 \\
\bottomrule
\end{tabular}
\end{table}

Table~\ref{tab:loss_ablation} fixes the \textbf{TIDES} architecture and removes one loss term at a time. The complete objective performs best across the primary metrics. Removing $L_{\mathrm{traj}}$ weakens transport accuracy, removing $L_{\mathrm{int}}$ degrades interaction-sensitive error and stabilization, and removing $L_{\mathrm{roll}}$ causes the clearest overall deterioration, especially on $l_{\mathrm{roll}}$ and $\mathrm{far}$. Exact implementation-level loss decompositions and weights are listed in Appendix~\ref{app:loss-details}.

\subsection{OOD and Long-Horizon Forecasting}
\label{sec:ood}

We evaluate two distribution shifts and one horizon extension. Motion/trajectory OOD uses unseen combinations of path length, speed, acceleration/deceleration, and straight/curved trajectories. Size OOD doubles sphere and cone radius and cube edge length at test time while keeping all models trained on the original dense benchmark. Horizon extension evaluates models trained at rollout length 8 at rollout lengths 16 and 24. Tables~\ref{tab:ood_motion}, \ref{tab:ood_shape}, and \ref{tab:ood_horizon} in appendix ~\ref{app:additional_ood_results} report these results.

The OOD and horizon results are consistent with the main comparison. \textbf{TIDES} gives the best aggregate behavior on motion OOD, size OOD, and both longer horizons, with the clearest stabilization diagnostic. \textbf{TIDE} remains competitive as the localized core, while external baselines can match isolated metrics but do not match the combined rollout, transport, and stabilization behavior. These results suggest that transport-aware coupling and explicit stabilization become most useful as distribution shift or accumulated rollout error increases.



\section{Discussion}

Our results support a local-to-global view of moving-solid-conditioned liquid forecasting: the interaction is initiated at the evolving solid--liquid interface, but the target is a full-domain Eulerian rollout. This is consistent with classical and learned FSI perspectives, where solid--fluid exchange is mediated through moving interfaces or boundary conditions~\citep{kolahdouz2021sharp,zhong2025benchmarking,tao2026neural}. The experiments indicate that localized interface modeling provides the core forecasting mechanism, while transport-aware feedback and explicit stabilization improve robustness under autoregressive rollout, OOD motion/size shifts, and horizon extension~\citep{koehler2024apebench,yagoubi2025neurips,hu2026realpdebench}.

\section{Limitations}

This benchmark should be interpreted as a controlled study rather than a replacement for physical simulation. The data are synthetic, the OOD axes are limited to motion and object-size shifts, and the dense Eulerian target is constructed from particle simulations rather than measured as a solver-native continuum field. Similar sim-to-real and benchmark-coverage limitations have been discussed in recent scientific ML benchmarks~\citep{hu2026realpdebench,yang2025multiphysics}. Future work should extend the task to richer geometries, broader physical parameter shifts, multi-solid interactions, and more practical repeated-query liquid forecasting.

\section{Conclusion}

We formulated \emph{interactive 4D volumetric liquid forecasting} as a moving-solid-conditioned dense prediction task: given a current liquid state and a known future solid motion, the model forecasts the future full-domain Eulerian liquid velocity rollout.
Across the controlled benchmark, our results identify localized interface-driven interaction as the central modeling principle, and show that transport-aware coupling and explicit stabilization improve long-horizon autoregressive forecasting.
We hope this task, benchmark, and TIDE/TIDES study provide a foundation for broader research on motion-conditioned liquid forecasting under richer geometries, broader distribution shifts, and more practical repeated-query settings.

\begin{ack}
This work was partly supported by the Institute of Information \& Communications Technology Planning \& Evaluation (IITP) grant funded by the Korea government (MSIT) [RS-2021-II211341, Artificial Intelligence Graduate School Program (ChungAng University)] and partly supported by the “Advanced GPU Utilization Support Program” funded by the Government of the Republic of Korea (Ministry of Science and ICT).
\end{ack}


\bibliographystyle{plainnat}
\bibliography{reference}

\clearpage

\appendix

\section{Adaptation Details for External Baselines}
\label{app:baseline_adapt}

All external baselines are trained against the same dense voxelized forecasting target used in the main benchmark: next-step Eulerian liquid velocity on the shared grid. The two voxel-space baselines, \textbf{3D ResUNet} and \textbf{U-NO}, receive the current liquid velocity together with global grid-aligned solid-conditioning channels derived from the future solid state. These channels encode solid-state cues such as boundary velocity and interface proximity, but the models do not use the localized mediator pathway, transport-feedback pathway, or explicit stabilization modules of the finalized method.

For \textbf{Transolver}, we use the same voxelized liquid input but concatenate a global solid-conditioning grid composed of voxelized boundary velocity and fluid--solid interface channels before applying full-grid physics-attention. For \textbf{GNS}, we preserve native Lagrangian particle prediction by representing the moving solid as kinematic rigid proxy particles and compute voxel-space metrics only after rasterizing predicted particle velocities onto the shared Eulerian grid. Unless otherwise noted, the same dense target-construction protocol is used for all methods so that every baseline is scored in the same prediction space.

Table~\ref{tab:baseline_adaptation_details} summarizes the input, conditioning, output, and evaluation space used for each adapted external baseline.

\begin{table}[t]
\centering
\caption{Implementation details for adapted external baselines under the dense-liquid, simplified-input protocol.}
\label{tab:baseline_adaptation_details}
\begin{tabular}{lllll}
\toprule
Model & Input & Conditioning & Output & Evaluation \\
\midrule
3D ResUNet & voxel grid & global solid channels & voxel velocity &  voxel \\
U-NO & voxel grid & global solid channels & voxel velocity & voxel \\
Transolver & voxel grid & global solid-conditioning grid & voxel velocity &  voxel \\
GNS & particle state & kinematic rigid proxies & particle state &  voxelized via P2G \\
\bottomrule
\end{tabular}
\end{table}

\section{Simulator Physics Behind Dataset Generation}
\label{app:simulator_physics}

The reported dense-liquid benchmark is generated from \textbf{DualSPHysics 5.4}, a weakly compressible smoothed particle hydrodynamics (WCSPH) Navier--Stokes solver for free-surface flow and fluid-driven rigid bodies. In our pipeline, case geometry is discretized by \texttt{GenCase}, the fluid simulation itself is executed by \texttt{DualSPHysics5.4}, fluid particles are exported by \texttt{PartVTK}, and rigid-body motion is exported by \texttt{FloatingInfo}. The resulting particle trajectories are then converted into the shared dense Eulerian benchmark target described in Appendix~\ref{app:metrics_p2g}.

At the continuum level, the solver advances a Lagrangian free-surface flow model governed by mass and momentum conservation,
\[
\frac{d\rho}{dt} = -\rho \nabla \cdot u,
\qquad
\frac{du}{dt} = -\frac{1}{\rho}\nabla p + g + \Gamma,
\]
where $\rho$ is density, $u$ velocity, $p$ pressure, $g$ gravity, and $\Gamma$ denotes dissipative terms such as viscosity. DualSPHysics discretizes these equations with SPH particle summations over a compact-support kernel. Pressure is obtained from a weakly compressible equation of state of Tait type,
\[
p = B\left[\left(\frac{\rho}{\rho_0}\right)^\gamma - 1\right],
\]
so that the simulation avoids a global pressure Poisson solve while keeping density fluctuations small when the artificial sound speed is sufficiently high. In the reported rigid-object cases, the case XML sets $\rho_0=1000\,\mathrm{kg/m^3}$, $\gamma=7$, gravity $g=(0,0,-9.81)$, \texttt{coefsound}=20, \texttt{hdp}=1.3, and \texttt{cflnumber}=0.2.

The reported runs use the following DualSPHysics numerical ingredients. The smoothing kernel is the \emph{Quintic Wendland} kernel. Dissipation is handled with \emph{Monaghan artificial viscosity}. Density noise is reduced with the \emph{Fourtakas} density-diffusion term. Particle disorder is controlled with the \emph{Lind et al.} shifting algorithm. Time integration uses the \emph{symplectic position-Verlet} scheme with a CFL-controlled variable timestep. Rigid-body motion is treated through the built-in floating-body formulation, in which pressure and viscous forces on boundary particles are accumulated into net force and torque and then integrated with rigid-body dynamics. In our reported rigid-object benchmark, the object is therefore not replayed as a purely kinematic animation; it remains fluid-coupled through the floating-body solver, while its nominal forcing schedule is generated by the runtime case-preparation script.

One implementation detail matters for interpreting boundary accuracy. The template XML is authored with \texttt{Boundary=2}, i.e.\ the modified dynamic boundary condition (mDBC), but the runtime case script switches the reported rigid closed-tank runs to \texttt{Boundary=1}, i.e.\ the classic dynamic boundary condition (DBC), because mDBC normal generation can fail in this enclosed floating-body setup. Thus, the benchmark should be understood as using the standard DualSPHysics WCSPH free-surface formulation together with a robust DBC rigid-wall treatment for the finalized rigid-object dataset. This is still a physically coupled CFD simulation, but it is not the most aggressive boundary-accuracy setting available in the general solver.

The benchmark input contract should be distinguished from the simulator internals. During simulation, the rigid object follows a realized fluid-coupled rigid-body trajectory under the runtime forcing schedule. After simulation, this realized future solid-state sequence is recorded and provided to the forecasting model as known conditioning. The model therefore receives future solid states but never future liquid states. In this sense, the benchmark assumes a known future moving-solid sequence and evaluates only the problem of forecasting the liquid response to that sequence.

We therefore treat the simulator output as a high-fidelity \emph{reference} rather than as exact ground truth. This is justified because DualSPHysics is a mature research and engineering code with extensive validations against analytical solutions, numerical benchmarks, and experimental data, and because WCSPH is particularly well matched to violent free-surface flow, splashing, and moving solid interaction. At the same time, the reference remains a numerical approximation: it inherits the assumptions of weak compressibility, artificial dissipation, boundary modelling, particle shifting, and finite resolution. In addition, our benchmark applies a separate particle-to-voxel target-construction stage, and some dataset-generation variants optionally add low-amplitude ambient surface perturbations during post-processing. For transparency, the dataset contract stores both raw and post-processed velocity fields when such perturbations are used, so that solver-native and augmented signals remain distinguishable.

\section{Efficiency and Reference-Simulator Latency}
\label{app:efficiency_latency}

Table~\ref{tab:efficiency_latency} reports the rollout-8 inference latency of the learned models together with a reference DualSPHysics timing derived from two representative benchmark logs. We use the simulator's reported \emph{simulation runtime per physical second} as the reference quantity and compare it against model inference for an 8-step rollout, since the finalized benchmark is sampled at 8 FPS. This comparison should be interpreted only as a controlled repeated-query latency reference under the benchmark contract, not as a claim that learned forecasting is equivalent to full CFD simulation.

\begin{table}[t]
\centering
\caption{Controlled rollout-8 latency comparison. Learned-model times are measured for a single 8-step query under the shared benchmark protocol. The DualSPHysics reference is the median simulation runtime per physical second across two representative benchmark logs.}
\label{tab:efficiency_latency}
\begin{tabular}{lcc}
\toprule
Method & Rollout-8 latency & Relative speed vs.\ DualSPHysics \\
\midrule
DualSPHysics (reference simulator) & 6.46\,s & 1.0$\times$ \\
\midrule
3D ResUNet & 130.2\,ms & 49.6$\times$ \\
U-NO & 201.9\,ms & 32.0$\times$ \\
GNS & 77.4\,ms & 83.5$\times$ \\
Transolver & 313.5\,ms & 20.6$\times$ \\
TIDE & 202.0\,ms & 32.0$\times$ \\
TIDES & 302.9\,ms & 21.3$\times$ \\
\bottomrule
\end{tabular}
\end{table}

The table makes two points relevant to the paper's positioning. First, \textbf{TIDES} remains substantially faster per query than the reference simulator under the benchmark protocol. Second, the more structured transport-aware models incur a meaningful latency cost relative to simpler voxel baselines, so the benchmark should be read as exposing an accuracy--latency tradeoff rather than as identifying a universally dominant surrogate.

\section{Exact Metric Definitions and Particle-to-Grid Conversion}
\label{app:metrics_p2g}

This section records the exact metric definitions and dense target-construction rules used in the reported experiments.

\paragraph{Rollout field error.}
Let $\hat{u}_{t}$ and $u_t$ denote the predicted and ground-truth Eulerian velocity fields at rollout step $t$. We first compute a per-step field error
\[
e_t =
\begin{cases}
\mathrm{mean}_{c,d,h,w}\,\ell(\hat{u}_t - u_t), & \text{without spatial weighting},\\[4pt]
\dfrac{\sum \ell(\hat{u}_t-u_t)\,w_{\mathrm{spatial}}}{\sum w_{\mathrm{spatial}}}, & \text{with spatial weighting},
\end{cases}
\]
where $\ell(x)=|x|$ or $x^2$ depending on the training configuration. The rollout metric is then
\[
l_{\mathrm{roll}} = \frac{1}{B}\sum_{b=1}^{B}
\frac{\sum_{t=1}^{T} \gamma^{t-1} e_t^{(b)}}{\sum_{t=1}^{T} \gamma^{t-1}}.
\]
In the current setting, $\gamma=1$, so $l_{\mathrm{roll}}$ reduces to the step-wise average rollout field error.

\paragraph{Near/Far velocity error.}
Let $S = |\phi^s|$ denote the absolute solid-distance field and let $M$ denote the valid dense liquid-domain mask. In the dense-liquid setting, $M$ is the full non-solid liquid domain rather than the sparse particle-support mask. The default near/far masks are defined by
\[
M_{\mathrm{near}} = \mathbf{1}[S \le \tau_{\mathrm{near}}] \cdot M,
\qquad
M_{\mathrm{far}} = \mathbf{1}[S \ge \tau_{\mathrm{far}}] \cdot M,
\]
with default thresholds $\tau_{\mathrm{near}} = 0.2$ and $\tau_{\mathrm{far}} = 0.9$.
If these masks become degenerate or overlap excessively, we fall back to quantile-based thresholds
\[
\tau_n = Q_{0.25}(S[M]), \qquad \tau_f = Q_{0.75}(S[M]),
\]
and define masks accordingly. If this still fails, we apply a rank-based split over valid voxels, assigning the lowest 25\% of interface distances to \emph{near} and the highest 25\% to \emph{far}. The corresponding metrics are masked velocity errors:
\[
\mathrm{near} = \mathrm{Err}(\hat{u},u;M_{\mathrm{near}}), \qquad
\mathrm{far} = \mathrm{Err}(\hat{u},u;M_{\mathrm{far}}).
\]

\paragraph{Trajectory endpoint error.}
Let $\hat{p}_{T}$ be particles advected by the predicted velocity field and $p_T$ the ground-truth particle endpoints. We define
\[
traj_{\mathrm{end}} = \frac{1}{N}\sum_{i=1}^{N}\|\hat{p}_{T,i} - p_{T,i}\|_2.
\]

\paragraph{Divergence delta.}
Let $u_{\mathrm{stage1\_pre}}$ denote the velocity before the final stabilization / projection stage and let $u_{\mathrm{final}}$ denote the final projected output. With
\[
\mathrm{DivScore}(u) = \mathrm{mean}\bigl(|\nabla \cdot u|\bigr),
\]
we define
\[
\Delta_{\mathrm{div}} = \mathrm{DivScore}(u_{\mathrm{final}}) - \mathrm{DivScore}(u_{\mathrm{stage1\_pre}}).
\]
Thus, more negative values indicate larger divergence reduction and therefore better stabilization.

\paragraph{Dense target construction from particles.}
Dataset construction begins with fluid-particle trajectories from the simulator. Let $p_i$ and $v_i$ denote particle positions and velocities. We first voxelize particles using trilinear/CIC splatting onto the Eulerian grid:
\[
W_g = \sum_i w_{ig}, \qquad
V_g = \sum_i w_{ig} v_i,
\]
and define a sparse particle-supported velocity field by
\[
u_g^{\mathrm{sparse}} =
\begin{cases}
V_g / W_g, & W_g > \epsilon,\\
0, & \text{otherwise}.
\end{cases}
\]
This yields a sparse support mask
\[
M_{\mathrm{supp}} = \mathbf{1}[W_g > \epsilon].
\]

In the dense-liquid setting, the final training target is not this sparse support field. Instead, we define the valid dense liquid domain as the full non-solid region
\[
M_{\mathrm{liq}} = 1 - M_{\mathrm{solid}},
\]
optionally clipped by a fixed water-level ceiling when constructing the benchmark. We then extend the sparse field over the valid dense liquid domain using nearest-support interpolation:
\[
u_g^{\mathrm{dense}} = u_{\pi(g)}^{\mathrm{sparse}},
\qquad
\pi(g)=\arg\min_{g' \in M_{\mathrm{supp}}}\|g-g'\|_2,
\qquad
g \in M_{\mathrm{liq}}.
\]
Outside the valid liquid domain, the target velocity is set to zero. Thus, the dataset stores both the sparse particle-support mask and the final dense target field, but the reported models are trained on the dense full-domain Eulerian target. This field should be read as a benchmark-defined dense Eulerian forecasting target constructed from particle simulations, not as an exact solver-native continuum field.

\paragraph{Different P2G operators used in the system.}
The dataset voxelizer above should not be confused with the internal dual-space transport feedback operator or the GNS evaluation voxelization. The dataset target constructor uses CIC splatting followed by dense-domain filling; the internal dual-space P2G uses nearest-voxel scatter of particle displacements; and the GNS evaluation pipeline uses a separate voxelization routine for converting predicted particle states into the shared Eulerian evaluation space.

\section{Implementation Details of the Reported Dense-Simplified Configuration}
\label{app:full-model-details}

\subsection{Operator Types and Tensor Shapes}
\label{app:operator-details}

The finalized dense-simplified configuration takes the current liquid velocity field together with a known future solid-state sequence. In the reported setting, the dense fluid tensor is
\[
x_t \in \mathbb{R}^{B \times 3 \times D \times H \times W},
\qquad (D,H,W) = (32,128,128),
\]
containing only the current Eulerian liquid velocity components.

The solid sequence is
\[
S_{t+1:t+T}
=
\{(\phi_{t+1}^s, v_{t+1}^s),\dots,(\phi_{t+T}^s, v_{t+T}^s)\},
\]
where
\[
\phi_\tau^s \in \mathbb{R}^{B \times 1 \times D \times H \times W},
\qquad
v_\tau^s \in \mathbb{R}^{B \times 3 \times D \times H \times W}.
\]
From each solid state, the implementation derives a 10-channel interface-aligned tensor consisting of fluid-solid interface, liquid-liquid interface, interface sdf, interface normal, boundary velocity, and boundary mask,
where the liquid--liquid interface channel is zero in the dense-liquid setting and is retained only for channel compatibility.

Legacy object-centric descriptors and control tensors remain available in the codebase for backward compatibility and exploratory ablations, but they are not part of the reported simplified input API.

The auxiliary particle state used in the Full Stabilized model is
\[
p_t \in \mathbb{R}^{B \times P \times 3},
\qquad P=256.
\]
When the explicit dual-space branch is enabled, the predicted particle trajectory tensor has shape
\[
\hat{p}_{1:T} \in \mathbb{R}^{B \times (T+1) \times P \times 3}.
\]

The fluid backbone is implemented as a 3-stage \texttt{UNOOperator3D}. Each stage uses a spectral-local residual block consisting of a 3D spectral convolution, a pointwise $1\times1\times1$ convolution, a local residual 3D convolution block, group normalization, and GELU activation.

The interface branch is implemented by \texttt{InterfaceBranchEncoder3D}. In the reported setting, it receives 10 interface channels, applies a 3D convolutional stem, and tokenizes the result through adaptive pooling and linear token projection. The interface token grid has spatial resolution $(2,4,4)$, yielding 32 global interface tokens in the default configuration. In the reported \textbf{TIDES} configuration, the simplified input pathway does not require a separate object-patch token. Object-local patch context appears only in the \textbf{TIDE} localized-family variants used in the localized-principle ablation.

The rollout core is implemented so that solid influence is mediated through the derived interface pathway rather than through naive global injection.

\subsection{Interaction Operator and One-Step Coupling Structure}
\label{app:interaction-details}

The finalized full model uses a structured interface-mediated interaction operator rather than naive global conditioning. At each step, the next-step solid state is first converted into a derived interface representation, which is then encoded into interface tokens. Let $z_{\mathrm{int}}^t$ denote the interface tokens and $z_f^t$ the fluid latent grid. The implemented update can be summarized as
\[
\mathcal{I}_{t+1} = \mathcal{D}(\phi_{t+1}^s, v_{t+1}^s),
\qquad
\tilde z_{\mathrm{int}}^{t+1} = E_{\mathrm{int}}(\mathcal{I}_{t+1}),
\]
\[
z_{\mathrm{int}}^{t+1}
=
z_{\mathrm{int}}^t
+ \mathrm{Attn}_{\mathrm{int}\leftarrow f}(z_f^t)
+ \tilde z_{\mathrm{int}}^{t+1}.
\]
The mediator pathway then performs interface-to-fluid exchange:
\[
\Delta z_{\mathrm{int}\rightarrow f}^t
=
\mathrm{Attn}_{f\leftarrow \mathrm{int}}(z_{\mathrm{int}}^{t+1}),
\]
which is reconstructed back to a grid-aligned latent residual using a token-to-grid path:
\[
\Delta z_{\mathrm{int}}^t
=
\mathrm{ImpulseRefine}\!\left(
\mathrm{TokensToGrid}(\Delta z_{\mathrm{int}\rightarrow f}^t)
\right).
\]
A learned interaction-intensity head produces a multiplicative amplification factor,
\[
I_t = h_{\mathrm{intensity}}\!\left(|\Delta z_{\mathrm{int}}^t|\right),
\qquad
\Delta z_{\mathrm{int}}^t
\leftarrow
\Delta z_{\mathrm{int}}^t \odot (1 + I_t).
\]

Bulk latent transport is computed independently by the fluid U-NO backbone:
\[
\Delta z_{\mathrm{bulk}}^t = F_\theta(z_f^t).
\]
The combined latent update before correction is therefore
\[
\tilde z_f^{t+1}
=
z_f^t + \Delta z_{\mathrm{bulk}}^t + \kappa \Delta z_{\mathrm{int}}^t.
\]

Thus, in the reported \textbf{TIDES} model, solid influence is not injected as an independent global control signal; instead, future solid geometry and motion are routed through the derived interface representation and only then reconstructed as a localized grid-aligned latent impulse.

\subsection{Transport Coupling and Correction Modules}
\label{app:transport-details}

The dual-space coupling pathway uses a grid-to-particle (G2P) update followed by a particle-to-grid (P2G) feedback step. First, a velocity-like field is decoded from the corrected latent state and sampled at particle locations using trilinear interpolation:
\[
u_{\mathrm{like}}^t = D_v(z_{f,\mathrm{lat1}}^{t+1}),
\qquad
p_{t+1}^{(n)} = \mathrm{clip}\!\left(p_t^{(n)} + \mathcal{S}(u_{\mathrm{like}}^t, p_t^{(n)})\right),
\]
where $\mathcal{S}$ denotes trilinear sampling.

The updated particle positions are then rasterized back to the grid through nearest-voxel scatter. Let
\[
g_n = \mathrm{round}(p_{t+1}^{(n)}).
\]
The model accumulates a density grid and displacement-sum grid:
\[
\rho(g) = \sum_n \mathbf{1}[g_n = g],
\qquad
d_{\mathrm{sum}}(g) = \sum_n (p_{t+1}^{(n)} - p_t^{(n)})\mathbf{1}[g_n = g].
\]
The mean displacement grid is
\[
\bar d(g) = \frac{d_{\mathrm{sum}}(g)}{\max(\rho(g),1)},
\]
and the density grid is normalized by the maximum density within each sample:
\[
\tilde \rho(g) = \frac{\rho(g)}{\max_{g'} \rho(g')}.
\]

The dual feedback tensor is then constructed by concatenating normalized density, mean displacement, and interface context:
\[
z_{\mathrm{p2g}}^t
=
\mathrm{DualRefine}\!\left(
[\tilde \rho,\bar d,\mathrm{IntCtx}]
\right),
\]
where \texttt{DualRefine} is a learned 3D convolutional refinement block. This correction is added back to the fluid latent grid before the second latent correction stage.

\subsection{Latent and Output Stabilization Modules}
\label{app:stabilization-details}

The latent correction modules $\Pi_{\mathrm{lat},1}$ and $\Pi_{\mathrm{lat},2}$ are both implemented by \texttt{LatentPressureCorrectionOperator}. These are not hard Poisson solvers; rather, they are lightweight learned residual correction modules with pressure-like structure.

Given an input latent tensor $z$, the operator first injects a global pooled summary:
\[
z_{\mathrm{glob}} = z + W_g\!\left(\mathrm{AvgPool3D}(z)\right),
\]
where the pooled summary is broadcast back to the original spatial resolution through a learned $1\times1\times1$ transform.

A scalar correction potential is then predicted:
\[
\phi = W_{\phi,2}\!\left(\mathrm{GELU}\!\left(W_{\phi,1}(z_{\mathrm{glob}})\right)\right),
\qquad
\phi \in \mathbb{R}^{B\times 1 \times D \times H \times W}.
\]
Its finite-difference gradient is computed as
\[
\nabla \phi = [\Delta_x \phi,\Delta_y \phi,\Delta_z \phi],
\]
with forward finite differences and zero padding at the spatial boundary. The correction is mapped back to latent space by a learned $1\times1\times1$ transform:
\[
\mathrm{corr}(z) = W_\nabla(\nabla \phi),
\]
and the corrected latent tensor is
\[
\Pi_{\mathrm{lat}}(z) = z_{\mathrm{glob}} - \alpha \,\mathrm{corr}(z).
\]

The first latent correction stage is applied before dual-space P2G feedback,
\[
z_{f,\mathrm{lat1}}^{t+1} = \Pi_{\mathrm{lat},1}(\tilde z_f^{t+1}),
\]
and the second is applied after adding dual feedback,
\[
z_{f,\mathrm{lat2}}^{t+1}
=
\Pi_{\mathrm{lat},2}\!\left(z_{f,\mathrm{lat1}}^{t+1} + \beta_{\mathrm{p2g}} z_{\mathrm{p2g}}^t\right).
\]

The output correction module receives the tentative decoded velocity field together with derived interface and boundary cues:
\[
x_{\mathrm{out}}
=
[\tilde u_{t+1}, s_{\mathrm{int}}, m_{\mathrm{bnd}}, u_{\mathrm{bnd}}, I_t],
\]
for a total of 9 input channels. The correction network consists of
\[
\mathrm{Conv3D} \rightarrow \mathrm{GN} \rightarrow \mathrm{GELU}
\rightarrow
\mathrm{Conv3D} \rightarrow \mathrm{GN} \rightarrow \mathrm{GELU}
\rightarrow
\mathrm{Conv1\times1\times1},
\]
producing a velocity correction $\Delta u$. The corrected output is
\[
u_{t+1}^{\mathrm{corr}} = \tilde u_{t+1} + \beta_{\mathrm{out}}\Delta u,
\]
followed by hard boundary overwrite:
\[
\hat u_{t+1}
=
(1-m_{\mathrm{bnd}})\odot u_{t+1}^{\mathrm{corr}}
+
m_{\mathrm{bnd}}\odot u_{\mathrm{bnd}}.
\]
A small feedback term derived from $\hat u_{t+1}-\tilde u_{t+1}$ and the interaction-intensity field is also fed back to the latent state.

\subsection{Loss Decomposition and Weights}
\label{app:loss-details}

While the main text groups the training objective into functional categories, the dense-simplified implementation uses the loss bundle
\[
\begin{aligned}
\mathcal{L}_{\mathrm{impl}}
={}&
\lambda_{\mathrm{vel}}\mathcal{L}_{\mathrm{vel}}
+\lambda_{\mathrm{grad}}\mathcal{L}_{\mathrm{grad}}
+\lambda_{\mathrm{div}}\mathcal{L}_{\mathrm{div}}
+\lambda_{\mathrm{traj}}\mathcal{L}_{\mathrm{traj}}
+\lambda_{\mathrm{int}}\mathcal{L}_{\mathrm{int}} \\
&+
\lambda_{\mathrm{roll}}\mathcal{L}_{\mathrm{roll}}
+\lambda_{\mathrm{phase}}\mathcal{L}_{\mathrm{phase}}
+\lambda_{\mathrm{dual}}\mathcal{L}_{\mathrm{dual}}
+\lambda_{\mathrm{exchange}}\mathcal{L}_{\mathrm{exchange}}
+\lambda_{\mathrm{particle}}\mathcal{L}_{\mathrm{particle}} .
\end{aligned}
\]

In the default dense-simplified configuration,
\[
\lambda_{\mathrm{vel}}=1.0,\qquad
\lambda_{\mathrm{grad}}=0.2,\qquad
\lambda_{\mathrm{div}}=0.1,\qquad
\lambda_{\mathrm{traj}}=0.5,\qquad
\lambda_{\mathrm{int}}=0.2,
\]
\[
\lambda_{\mathrm{roll}}=0.5,\qquad
\lambda_{\mathrm{phase}}=0.1,\qquad
\lambda_{\mathrm{dual}}=0.5,\qquad
\lambda_{\mathrm{exchange}}=0.1,\qquad
\lambda_{\mathrm{particle}}=0.2.
\]

Table~\ref{tab:loss_weights} summarizes the implementation-level loss weights used by the dense-simplified full model.

\begin{table}[t]
\centering
\caption{Loss terms used in the dense-simplified implementation.}
\label{tab:loss_weights}
\begin{tabular}{lcc}
\toprule
Loss term & Symbol & Weight \\
\midrule
Velocity reconstruction & $\lambda_{\mathrm{vel}}$ & 1.0 \\
Gradient / vorticity consistency & $\lambda_{\mathrm{grad}}$ & 0.2 \\
Divergence regularization & $\lambda_{\mathrm{div}}$ & 0.1 \\
Trajectory supervision & $\lambda_{\mathrm{traj}}$ & 0.5 \\
Interface consistency & $\lambda_{\mathrm{int}}$ & 0.2 \\
Rollout consistency & $\lambda_{\mathrm{roll}}$ & 0.5 \\
Phase supervision & $\lambda_{\mathrm{phase}}$ & 0.1 \\
Dual-space consistency & $\lambda_{\mathrm{dual}}$ & 0.5 \\
Mediator exchange balance & $\lambda_{\mathrm{exchange}}$ & 0.1 \\
Explicit particle-branch supervision & $\lambda_{\mathrm{particle}}$ & 0.2 \\
\bottomrule
\end{tabular}
\end{table}

Here, $\mathcal{L}_{\mathrm{vel}}$ is the one-step dense Eulerian reconstruction loss, $\mathcal{L}_{\mathrm{grad}}$ matches spatial velocity derivatives and optional vorticity, $\mathcal{L}_{\mathrm{div}}$ is the mean absolute divergence regularizer on the predicted rollout, $\mathcal{L}_{\mathrm{traj}}$ supervises particles obtained by advecting sampled particles through the predicted Eulerian field, and $\mathcal{L}_{\mathrm{roll}}$ is the multi-step weighted dense rollout error. In the reported benchmark, the phase term should be read narrowly: it is an auxiliary two-channel liquid/solid occupancy supervision used in the full-model family, rather than a general multi-phase prediction objective used by every baseline. Simpler localized and external baselines disable this head.

The interface loss is
\[
\mathcal{L}_{\mathrm{int}}
=
w_{\mathrm{bv}}\mathcal{L}_{\mathrm{boundary\_velocity}}
+
w_{\mathrm{np}}\mathcal{L}_{\mathrm{no\_penetration}}
+
w_{\mathrm{n}}\mathcal{L}_{\mathrm{normal\_consistency}}.
\]
The dual-space consistency term $\mathcal{L}_{\mathrm{dual}}$ measures Eulerian--Lagrangian consistency under the same sampled particles, $\mathcal{L}_{\mathrm{exchange}}$ penalizes imbalance in the mediator exchange pathway, and $\mathcal{L}_{\mathrm{particle}}$ supervises the explicit particle rollout branch when present.

For presentation in the main paper, these implementation-level terms are grouped into the five functional categories of field reconstruction, rollout consistency, interface consistency, transport consistency, and stabilization. Simpler baselines and ablations disable the terms that are irrelevant to their architecture.

\section{Seed Stability of the Main Comparison}
\label{app:seed_stability}

Table~\ref{tab:main_comparison_std} reports the same main-comparison means as Table~\ref{tab:main_comparison}, together with standard deviations across 5 random seeds. We include this table to make the seed variability explicit while keeping the main text focused on the aggregate comparison.

\begin{table}[b]
\centering
\caption{Seed-stability version of Table~\ref{tab:main_comparison}. The mean values match the main-text table, and the appendix additionally reports mean $\pm$ standard deviation across 5 random seeds. Lower is better.}

\label{tab:main_comparison_std}
\begin{tabular}{lccccc}
\toprule
Model & $l_{\mathrm{roll}}\downarrow$ & $\mathrm{traj\_end}\downarrow$ & $\mathrm{near}\downarrow$ & $\mathrm{far}\downarrow$ & $\Delta_{\mathrm{div}}$ \\
\midrule

3D ResUNet  &
\makecell[c]{$0.007585$ \\ $\pm 0.0011$} &
\makecell[c]{$0.122781$ \\ $\pm 0.1412$} &
\makecell[c]{$0.005757$ \\ $\pm 0.0015$} &
\makecell[c]{$0.009553$ \\ $\pm 0.0031$} &
\makecell[c]{$+0.000000$ \\ $\pm 0.0007$} \\

U-NO  &
\makecell[c]{$0.006542$ \\ $\pm 0.0005$} &
\makecell[c]{$0.112111$ \\ $\pm 0.1679$} &
\makecell[c]{$0.005807$ \\ $\pm 0.0020$} &
\makecell[c]{$0.005999$ \\ $\pm 0.0021$} &
\makecell[c]{$-0.000376$ \\ $\pm 0.0008$} \\

GNS  &
\makecell[c]{$3.767170$ \\ $\pm 0.0286$} &
\makecell[c]{$5.555210$ \\ $\pm 1.8474$} &
\makecell[c]{$3.673578$ \\ $\pm 0.0286$} &
\makecell[c]{$3.957512$ \\ $\pm 0.0286$} &
\makecell[c]{$+0.000000$ \\ $\pm 0.0032$} \\

Transolver  &
\makecell[c]{$0.006981$ \\ $\pm 0.0007$} &
\makecell[c]{$0.142872$ \\ $\pm 0.0575$} &
\makecell[c]{$0.005712$ \\ $\pm 0.0024$} &
\makecell[c]{$0.008737$ \\ $\pm 0.0028$} &
\makecell[c]{$+0.000000$ \\ $\pm 0.0007$} \\

\midrule

Bulk Only &
\makecell[c]{$0.007366$ \\ $\pm 0.0007$} &
\makecell[c]{$0.121042$ \\ $\pm 0.0188$} &
\makecell[c]{$0.005771$ \\ $\pm 0.0018$} &
\makecell[c]{$0.007495$ \\ $\pm 0.0020$} &
\makecell[c]{$+0.000241$ \\ $\pm 0.0007$} \\

Finite Support &
\makecell[c]{$0.006240$ \\ $\pm 0.0009$} &
\makecell[c]{$0.113406$ \\ $\pm 0.0413$} &
\makecell[c]{$0.005855$ \\ $\pm 0.0019$} &
\makecell[c]{$0.006389$ \\ $\pm 0.0022$} &
\makecell[c]{$-0.000150$ \\ $\pm 0.0007$} \\

TIDE &
\makecell[c]{$0.006113$ \\ $\pm 0.0005$} &
\makecell[c]{$0.110904$ \\ $\pm 0.0048$} &
\makecell[c]{$0.005660$ \\ $\pm 0.0014$} &
\makecell[c]{$0.006279$ \\ $\pm 0.0020$} &
\makecell[c]{$-0.000248$ \\ $\pm 0.0009$} \\

TIDES &
\makecell[c]{$\mathbf{0.005746}$ \\ $\mathbf{\pm 0.0005}$} &
\makecell[c]{$\mathbf{0.104397}$ \\ $\mathbf{\pm 0.0036}$} &
\makecell[c]{$\mathbf{0.005581}$ \\ $\mathbf{\pm 0.0013}$} &
\makecell[c]{$\mathbf{0.005826}$ \\ $\mathbf{\pm 0.0018}$} &
\makecell[c]{$\mathbf{-0.057067}$ \\ $\mathbf{\pm 0.0114}$} \\
\bottomrule
\end{tabular}
\end{table}

\section{Additional OOD and Long-Horizon Results}
\label{app:additional_ood_results}

Tables~\ref{tab:ood_motion}, \ref{tab:ood_shape}, and \ref{tab:ood_horizon} report the numeric OOD and horizon-extension results referenced in Section~\ref{sec:ood}. We place these tables in the appendix to keep the main experimental narrative compact while preserving the full metric breakdown.

\begin{table}[t]
\centering
\caption{OOD results on motion/trajectory OOD. Lower is better.}
\label{tab:ood_motion}
\begin{tabular}{lccccc}
\toprule
Model & $l_{\mathrm{roll}}\downarrow$ & $traj_{\mathrm{end}}\downarrow$ & $\mathrm{near}\downarrow$ & $\mathrm{far}\downarrow$ & $\Delta_{\mathrm{div}}$ \\
\midrule
3D ResUNet & 0.008630 & 0.135718 & 0.006614 & 0.010646 & +0.000000 \\
U-NO  & 0.007816 & 0.140968 & 0.006769 & 0.007073 & -0.000371 \\
Transolver & 0.008277 & 0.211586 & 0.006864 & 0.010068 & +0.000000 \\
GNS  & 4.150108 & 8.790151 & 4.150108 & 4.150108 & +0.000000 \\
\midrule
TIDE & 0.007153 & 0.146672 & 0.006494 & 0.007417 & -0.000224 \\
TIDES & \textbf{0.006771} & \textbf{0.132804} & \textbf{0.006457} & \textbf{0.006950} & \textbf{-0.056732} \\
\bottomrule
\end{tabular}
\end{table}

\begin{table}[t]
\centering
\caption{OOD results on official size OOD, where held-out sphere and cone cases use doubled radius (2x) and held-out cube cases use doubled edge length (2x) at test time. Lower is better.}
\label{tab:ood_shape}
\begin{tabular}{lccccc}
\toprule
Model & $l_{\mathrm{roll}}\downarrow$ & $traj_{\mathrm{end}}\downarrow$ & $\mathrm{near}\downarrow$ & $\mathrm{far}\downarrow$ & $\Delta_{\mathrm{div}}$ \\
\midrule

3D ResUNet & 0.021598 & 2.264266 & 0.011117 & 0.029733 & +0.000000 \\
U-NO & 0.005987 & 0.698487 & 0.005981 & 0.006846 & -0.000364 \\
Transolver & 0.009305 & 0.843196 & 0.009604 & 0.121690 & +0.000000 \\
GNS  & 3.107471 & 4.115864 & 3.085135 & 3.117987 & +0.000000 \\
\midrule
Bulk Only & 0.006530 & 0.773276 & 0.005943 & 0.006590 & +0.000214 \\

Finite Support & 0.006786 & 0.715258 & 0.006279 & 0.006669 & -0.000159 \\
TIDE & 0.006519 & 0.690246 & 0.005904 & 0.006485 & -0.000220 \\
TIDES & \textbf{0.006018} & \textbf{0.684433} & \textbf{0.005857} & \textbf{0.006022} & \textbf{-0.056036} \\
\bottomrule
\end{tabular}
\end{table}

\begin{table}[t]
\centering
\caption{Long-horizon forecasting on the in-distribution test split after training with horizon 8. Rows are grouped by evaluation rollout step. Lower is better.}
\label{tab:ood_horizon}
\begin{tabular}{clcccc}
\toprule
Step & Model & $l_{\mathrm{roll}}\downarrow$ & $\mathrm{near}\downarrow$ & $\mathrm{far}\downarrow$ & $\Delta_{\mathrm{div}}$ \\
\midrule
\textit{16} & 3D ResUNet  & 0.009947 & 0.006339 & 0.012917 & +0.000000 \\
& U-NO  & 0.006632 & 0.005664 & 0.005562 & -0.000296 \\
& GNS  & 8.067133 & 8.065323 & 8.277538 & +0.000000 \\
& Transolver & 0.009115 & 0.006001 & 0.011649 & +0.000000 \\
\cmidrule(lr){2-6}
& TIDE & 0.006366 & 0.006130 & 0.006269 & -0.000115 \\
& TIDES & \textbf{0.005964} & \textbf{0.005911} & \textbf{0.005869} & \textbf{-0.048668} \\
\midrule
\textit{24} & 3D ResUNet  & 0.011981 & 0.007319 & 0.015942 & +0.000000 \\
& U-NO  & 0.006826 & 0.006991 & 0.005877 & -0.000267 \\
& GNS  & 8.714556 & 8.704857 & 8.918174 & +0.000000 \\
& Transolver & 0.010734 & 0.006887 & 0.013558 & +0.000000 \\
\cmidrule(lr){2-6}
& TIDE & 0.006489 & 0.006513 & 0.006105 & -0.000071 \\
& TIDES & \textbf{0.006089} & \textbf{0.006219} & \textbf{0.005744} & \textbf{-0.045924} \\
\bottomrule
\end{tabular}
\end{table}

\begin{figure}[t]
    \centering
    \includegraphics[width=\linewidth]{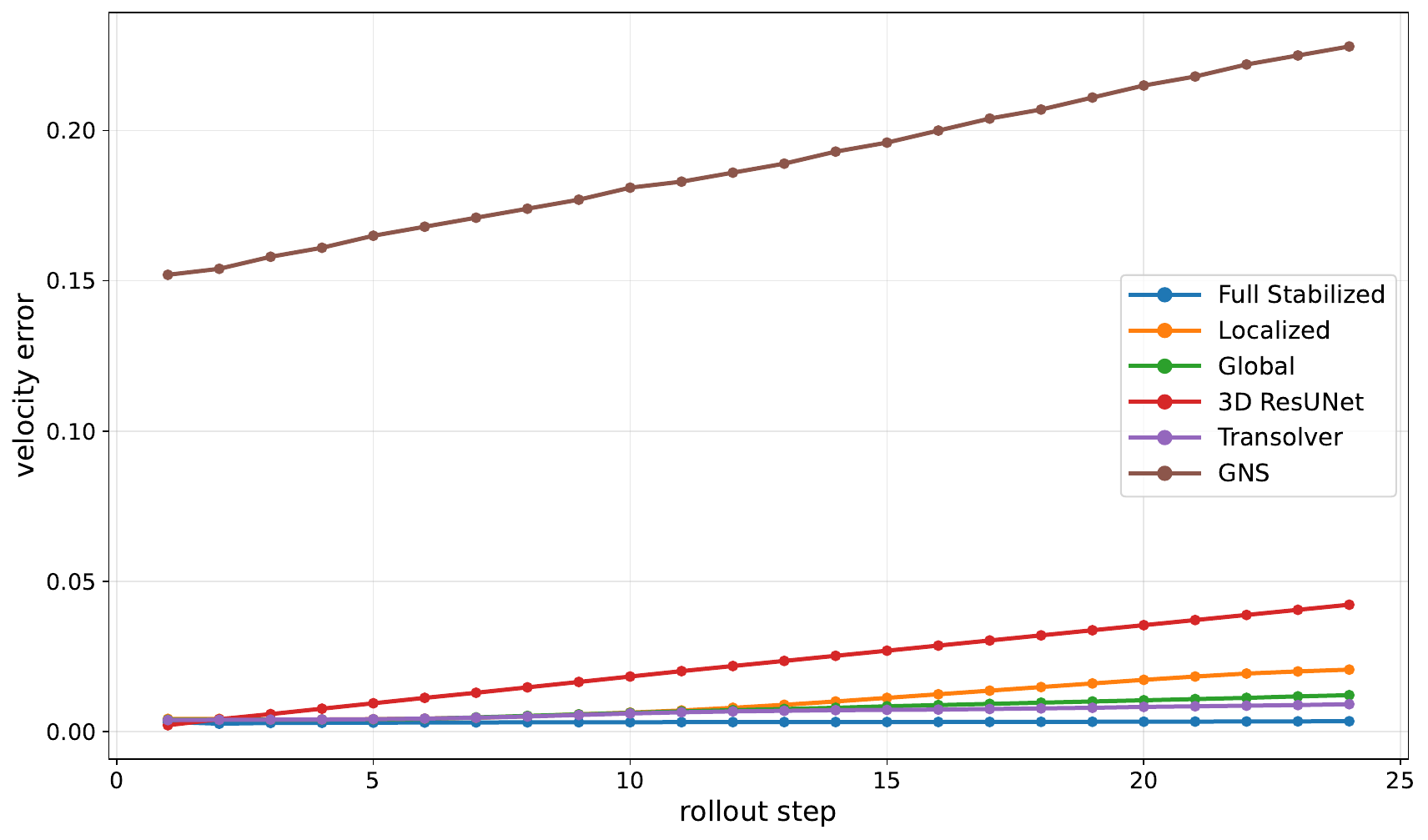}
    \caption{Long-horizon velocity error growth on the test split.}
    \label{fig:ood_horizon}
\end{figure}

\begin{figure}[t]
    \centering
    \includegraphics[width=\linewidth]{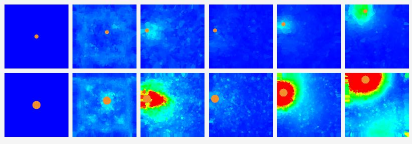}
    \caption{Representative size-OOD dataset examples used for object-scale generalization evaluation.}
    \label{fig:ood_dataset}
\end{figure}

Figure~\ref{fig:ood_horizon} visualizes long-horizon velocity-error growth, complementing the numeric results in Table~\ref{tab:ood_horizon}. Figure~\ref{fig:ood_dataset} shows representative size-OOD examples used to evaluate object-scale generalization.

\section{Additional Controlled Baseline Comparisons}
\label{app:matched_control}

\subsection{Parameter-Matched Localized U-NO Control}

To separate the effect of localized conditioning from the additional
TIDE/TIDES mechanisms, we introduce a direct-localized U-NO control.
It receives the same solid occupancy, boundary velocity, and derived
interface representation, but does not use the TIDE mediator or the
TIDES transport and correction modules.

This experiment uses the same train/validation/test splits, target
construction, normalization, rollout definition, and evaluator as
Table~\ref{tab:main_comparison}. However, it uses a separate training
protocol. The original main comparison uses a fixed optimization budget
across model families. For this additional control, we adjust the U-NO
width to match TIDE's approximately 338M-parameter budget and retrain
all four models with a common extended optimization schedule and five
paired random seeds. The longer schedule reduces the risk that the
modified U-NO configurations are disadvantaged by insufficient
optimization. TIDE and TIDES retain their original parameter counts
but are retrained under the same extended schedule.

Consequently, the means in Table~\ref{tab:matched_localized_control}
are comparable within this additional experiment and should be
distinguished from the original fixed-budget results in
Table~\ref{tab:main_comparison}.


\begin{table}[t]
\centering
\caption{Additional parameter-matched control under a common extended
optimization schedule. Values are means over five paired training
seeds. The final column reports rollout error at evaluation horizon 24.
Lower forecasting error is better.}
\label{tab:matched_localized_control}
\begin{tabular}{lcccc}
\toprule
Model & Params. & $l_{\mathrm{roll}}$
& $traj_{\mathrm{end}}$ & $l_{\mathrm{roll}}^{H=24}$ \\
\midrule
U-NO, global input & 338.0M & 0.00570 & 0.132 & 0.00683 \\
U-NO, localized input & 337.9M & 0.00548 & 0.124 & 0.00660 \\
TIDE & 338.0M & 0.00530 & 0.112 & 0.00638 \\
TIDES & 338.3M & \textbf{0.00502}
& \textbf{0.104} & \textbf{0.00598} \\
\bottomrule
\end{tabular}
\end{table}

Direct localized inputs improve U-NO's mean rollout error from
0.00570 to 0.00548. Relative to this localized control, TIDE reduces
mean rollout error by 3.3\% and trajectory endpoint error by 9.7\%.
TIDES further reduces mean rollout error by 5.3\% relative to TIDE.
These results indicate that localized inputs account for part of
the improvement, while the additional mediation and
transport/stabilization mechanisms provide further mean gains.

The absolute localized-U-NO-to-TIDE rollout difference is 0.00018.
We interpret this as a modest mean improvement and do not claim
statistical significance from the aggregate means alone. For paired
seed comparisons, the relevant quantity is
\[
d_j =
l_{\mathrm{roll},j}^{\mathrm{localized\ U\mbox{-}NO}}
-
l_{\mathrm{roll},j}^{\mathrm{TIDE}},
\qquad j=1,\ldots,5,
\]
where a positive $d_j$ indicates lower TIDE error for the same seed.


\subsection{Shared Boundary and Physical-Loss Controls}

We additionally test whether unequal access to boundary enforcement
or output-level physical losses explains the model ordering.
The same hard boundary overwrite is applied to U-NO, TIDE, and TIDES,
using the operator defined in Appendix~\ref{app:stabilization-details}.
A separate training control applies the same applicable output-level
physical losses across the three models.

Table~\ref{tab:shared_physical_controls} reports the resulting
rollout errors. Each additional setting improves the reported mean
error, while TIDES retains the lowest value. Under the shared
physical-loss setting, its error is 5.4\% below TIDE and 10.4\%
below U-NO.


\begin{table}[t]
\centering
\caption{Rollout error under shared boundary enforcement and
physical-loss controls. The reference column uses the additional
extended-schedule runs.}
\label{tab:shared_physical_controls}
\begin{tabular}{lccc}
\toprule
Model & Reference & Shared overwrite & Shared physical losses \\
\midrule
U-NO & 0.00570 & 0.00565 & 0.00549 \\
TIDE & 0.00530 & 0.00527 & 0.00520 \\
TIDES & \textbf{0.00502} & \textbf{0.00499}
& \textbf{0.00492} \\
\bottomrule
\end{tabular}
\end{table}

\section{Target and Solid-Conditioning Sensitivity}
\label{app:target_conditioning_sensitivity}

\subsection{Particle-to-Grid Target Sensitivity}

The dense target is constructed from particle simulations according
to Appendix~\ref{app:metrics_p2g}. To assess dependence on this
construction, we evaluate the same fixed model predictions against
alternative targets and spatial supports: nearest-support filling,
SPH-kernel interpolation, kernel-supported voxels only, and a
smoothed dense target.

Table~\ref{tab:target_construction_sensitivity} shows that the
absolute error scale changes, while the reported ordering remains
unchanged. TIDES has approximately 8.3--8.4\% lower error than
localized U-NO across these four evaluations.

Direct particle-hit voxels occupy 4.2\% of the evaluated liquid grid,
whereas 93.8\% of liquid voxels lie within the SPH kernel support of
at least one particle. The remaining 6.2\% lie outside that kernel
support. Direct particle hits, kernel support, and the CIC support
mask used for target construction are distinct support definitions
and should not be conflated.

These checks hold predictions fixed; they do not evaluate retraining
under alternative target constructions. The preserved ordering
reduces concern that nearest-support filling alone produces the
comparison, but does not establish that the constructed target is an
exact continuum velocity field.


\begin{table}[t]
\centering
\caption{Target-construction sensitivity using fixed predictions.
Lower rollout error is better.}
\label{tab:target_construction_sensitivity}
\begin{tabular}{lccc}
\toprule
Evaluation target or support & Localized U-NO & TIDE & TIDES \\
\midrule
Nearest-support dense target & 0.00548 & 0.00530
& \textbf{0.00502} \\
SPH-kernel interpolation & 0.00542 & 0.00524
& \textbf{0.00497} \\
Kernel-supported voxels only & 0.00535 & 0.00517
& \textbf{0.00490} \\
Smoothed dense target & 0.00527 & 0.00510
& \textbf{0.00483} \\
\bottomrule
\end{tabular}
\end{table}

\subsection{Sensitivity to Future Solid-State Errors}

We perturb the supplied future solid-state sequence with temporally
smooth trajectory deviations while retaining the original reference
liquid rollout. Perturbation magnitude is reported as the realized
RMS solid-position discrepancy in voxel units.
Table~\ref{tab:solid_conditioning_sensitivity} measures how inaccurate
conditioning affects TIDES.

Rollout error increases by 1.6\% at a 0.40-voxel RMS discrepancy and
13.3\% at 1.60 voxels. The result characterizes robustness to
conditioning error; it does not establish performance under
independently simulated prescribed trajectories or closed-loop
fluid--solid coupling. Here, $traj_{\mathrm{end}}$ measures forecast
liquid-trajectory endpoints, rather than the error in the solid
trajectory.


\begin{table}[t]
\centering
\caption{TIDES sensitivity to smooth perturbations of future
solid-state conditioning. The final column gives the relative
increase in rollout error over unperturbed conditioning.}
\label{tab:solid_conditioning_sensitivity}
\begin{tabular}{ccccc}
\toprule
RMS error (voxels) & $l_{\mathrm{roll}}$
& Interface error & $traj_{\mathrm{end}}$ & Increase \\
\midrule
0.00 & 0.00502 & 0.0254 & 0.104 & 0.0\% \\
0.16 & 0.00504 & 0.0256 & 0.105 & 0.4\% \\
0.40 & 0.00510 & 0.0262 & 0.108 & 1.6\% \\
0.80 & 0.00528 & 0.0280 & 0.114 & 5.2\% \\
1.60 & 0.00569 & 0.0327 & 0.129 & 13.3\% \\
\bottomrule
\end{tabular}
\end{table}

\section{Additional Temporal and Physical Diagnostics}
\label{app:temporal_physical_diagnostics}

The following evaluations use free-running autoregressive predictions
after training with an eight-step horizon. For notation, let $t$
denote the observed starting time and $k$ the forecast offset:
$k=1,\ldots,H$. Only the initial liquid state is observed; subsequent
liquid inputs come from the model's predictions, while the future
solid-state sequence remains supplied.

\subsection{Per-Step Interface and Far-Field Error}

We evaluate U-NO, TIDE, and TIDES on the same 64 in-distribution test
windows. To define regions independently of the filled velocity
target, let $O_{t+k}$ be the simulator solid-occupancy mask,
$M_{t+k}$ the valid liquid mask, and $\mathcal{D}_r$ voxel dilation
with radius $r$. The diagnostic regions are
\[
M_{t+k}^{\mathrm{int}}
=
M_{t+k}\cap\mathcal{D}_4(O_{t+k}),
\qquad
M_{t+k}^{\mathrm{far}}
=
M_{t+k}\setminus\mathcal{D}_8(O_{t+k}).
\]
The intermediate four-to-eight-voxel band is excluded from these two
diagnostics. These masks differ from the original distance-based
near/far definitions in Appendix~\ref{app:metrics_p2g}; the resulting
numbers should therefore be interpreted separately.

Table~\ref{tab:temporal_interface_far} reports selected points from
the 24-step evaluation. Errors need not increase monotonically
because motion magnitude and interface location vary over time.
At offset 24, TIDES reduces interface and far-field errors by 5.4\%
and 9.6\%, respectively, relative to TIDE.


\begin{table}[t]
\centering
\caption{Selected free-running interface/far-field errors.
Each cell reports interface error / far-field error under the
geometry-derived diagnostic masks.}
\label{tab:temporal_interface_far}
\begin{tabular}{lcccc}
\toprule
Model & $t+1$ & $t+8$ & $t+16$ & $t+24$ \\
\midrule
U-NO
& 0.0393 / 0.00594
& 0.0256 / 0.00660
& 0.0263 / 0.00595
& 0.0408 / 0.00650 \\
TIDE
& 0.0384 / 0.00590
& 0.0250 / 0.00635
& 0.0260 / 0.00574
& 0.0407 / 0.00625 \\
TIDES
& \textbf{0.0380 / 0.00585}
& \textbf{0.0245 / 0.00575}
& \textbf{0.0248 / 0.00520}
& \textbf{0.0385 / 0.00565} \\
\bottomrule
\end{tabular}
\end{table}

\subsection{Temporal-Change and Velocity-Derived Diagnostics}

Let $\langle\cdot\rangle_M$ denote averaging over the common valid
spatial mask, including component averaging for vector quantities.
We set $\hat{u}_t=u_t$ so that the first predicted temporal increment
is measured relative to the shared observed state. Temporal-change
error is
\[
E_{\mathrm{change}}
=
\frac{1}{H}\sum_{k=1}^{H}
\left\langle
\left|
(\hat{u}_{t+k}-\hat{u}_{t+k-1})
-
(u_{t+k}-u_{t+k-1})
\right|
\right\rangle_M .
\]
This is an error in frame-to-frame velocity increments.

Additional diagnostics compare divergence, vorticity, and kinetic
energy derived from the velocity fields:
\[
\begin{aligned}
E_{\mathrm{div}}
&=
\frac{1}{H}\sum_{k=1}^{H}
\left\langle
\left|
\nabla_g\cdot\hat{u}_{t+k}
-
\nabla_g\cdot u_{t+k}
\right|
\right\rangle_M, \\
D_{\mathrm{abs}}
&=
\frac{1}{H}\sum_{k=1}^{H}
\left\langle
\left|\nabla_g\cdot\hat{u}_{t+k}\right|
\right\rangle_M, \\
E_{\mathrm{vort}}
&=
\frac{1}{H}\sum_{k=1}^{H}
\left\langle
\left|
\nabla_g\times\hat{u}_{t+k}
-
\nabla_g\times u_{t+k}
\right|
\right\rangle_M, \\
E_{\mathrm{energy}}
&=
\frac{1}{H}\sum_{k=1}^{H}
\left\langle
\left|
q(\hat{u}_{t+k})-q(u_{t+k})
\right|
\right\rangle_M,
\qquad
q(u)=\frac{1}{2}\sum_{c=1}^{3}u_c^2 .
\end{aligned}
\]
Here, $\nabla_g$ denotes the common discrete spatial operator.
The energy diagnostic uses unit-density kinetic energy in the
evaluation coordinates. All models use identical masks and spatial
operators.

Table~\ref{tab:additional_velocity_diagnostics} reports means over
offsets 1--24. TIDES has the lowest reported value in all five columns.
In particular, its mean absolute divergence is 37.2\% below TIDE.
The accompanying vorticity and energy errors provide complementary
evidence when interpreting divergence reduction, although they do
not establish exact conservation or rule out every form of damping.

Unlike $\Delta_{\mathrm{div}}$, which measures a change across
correction stages, $D_{\mathrm{abs}}$ measures absolute divergence
of the final prediction. Neither quantity alone establishes physical
fidelity. Pressure and free-surface state are not common prediction
targets across the compared models, so these diagnostics do not
constitute a complete physical evaluation.

\begin{table}[t]
\centering
\caption{Additional diagnostics averaged over free-running
offsets 1--24. Lower values are better.}
\label{tab:additional_velocity_diagnostics}
\begin{tabular}{lccccc}
\toprule
Model & $E_{\mathrm{change}}$ & $E_{\mathrm{div}}$
& $D_{\mathrm{abs}}$ & $E_{\mathrm{vort}}$
& $E_{\mathrm{energy}}$ \\
\midrule
U-NO & 0.004086 & 0.003054 & 0.002511
& 0.002359 & 0.000573 \\
TIDE & 0.004169 & 0.004082 & 0.002517
& 0.002670 & 0.000610 \\
TIDES & \textbf{0.003960} & \textbf{0.003020}
& \textbf{0.001580} & \textbf{0.002330}
& \textbf{0.000565} \\
\bottomrule
\end{tabular}
\end{table}

\subsection{Free-Running Evaluation at Horizons 48 and 64}

Table~\ref{tab:extended_rollout_64} extends evaluation to horizons
48 and 64. It supplements the original horizon-16/24 results in
Table~\ref{tab:ood_horizon}; the original fixed-budget results remain
identified by their earlier protocol.

At horizon 64, TIDES has 11.0\% lower rollout error than TIDE and
18.4\% lower error than U-NO. Its absolute divergence at offset 64 is
36.4\% below TIDE and 44.7\% below U-NO. The measurements show gradual
error growth over the evaluated horizons, rather than abrupt failure.
They do not establish stability at arbitrarily long horizons.


\begin{table}[t]
\centering
\caption{Additional long-horizon evaluation. Rollout errors are
reported at evaluation horizons 48 and 64; divergence columns report
absolute predicted divergence at the corresponding final offsets.}
\label{tab:extended_rollout_64}
\begin{tabular}{lcccc}
\toprule
Model & $l_{\mathrm{roll}}^{H=48}$
& $l_{\mathrm{roll}}^{H=64}$
& $D_{\mathrm{abs},48}$ & $D_{\mathrm{abs},64}$ \\
\midrule
U-NO & 0.00755 & 0.00815 & 0.0033 & 0.0038 \\
TIDE & 0.00713 & 0.00747 & 0.0031 & 0.0033 \\
TIDES & \textbf{0.00642} & \textbf{0.00665}
& \textbf{0.0019} & \textbf{0.0021} \\
\bottomrule
\end{tabular}
\end{table}

Table~\ref{tab:spatial_divergence_64} further characterizes the
spatial distribution of absolute divergence at offset 64.
Both the regional mean and the 95th percentile are lower for TIDES
near the interface and in the far field. This provides a regional
view of stabilization beyond a single global average.

\begin{table}[t]
\centering
\caption{Regional absolute-divergence statistics at offset 64.
Each cell reports mean / 95th percentile.}
\label{tab:spatial_divergence_64}
\begin{tabular}{lcc}
\toprule
Model & Interface & Far field \\
\midrule
U-NO & 0.0050 / 0.0140 & 0.0026 / 0.0065 \\
TIDE & 0.0049 / 0.0155 & 0.0022 / 0.0061 \\
TIDES & \textbf{0.0031 / 0.0090}
& \textbf{0.0015 / 0.0040} \\
\bottomrule
\end{tabular}
\end{table}

\subsection{Advected-Tracer Volume Diagnostic}

To examine volume behavior under predicted transport, we advect a
common initial tracer set and reconstruct its occupied volume.
Every model uses identical tracer initialization, interpolation,
advection, and reconstruction. Table~\ref{tab:tracer_volume_drift}
reports the resulting relative volume drift.

TIDES reaches 2.8\% drift at offset 64, compared with 4.8\% for TIDE
and 5.5\% for U-NO. This diagnostic measures volume behavior induced
by velocity-based tracer transport. It depends on the shared
reconstruction procedure and is not a direct measurement of exact
mass conservation or predicted free-surface geometry.


\begin{table}[t]
\centering
\caption{Relative volume drift from shared advected-tracer
reconstruction. Lower drift is better.}
\label{tab:tracer_volume_drift}
\begin{tabular}{lccc}
\toprule
Model & Offset 24 & Offset 48 & Offset 64 \\
\midrule
U-NO & 2.5\% & 4.2\% & 5.5\% \\
TIDE & 2.3\% & 3.8\% & 4.8\% \\
TIDES & \textbf{1.4\%} & \textbf{2.1\%}
& \textbf{2.8\%} \\
\bottomrule
\end{tabular}
\end{table}

\section{Shape and Rotation Breakdown}
\label{app:shape_rotation_breakdown}

We disaggregate the additional evaluation by object family and
solid angular motion. Table~\ref{tab:shape_rotation_breakdown}
reports sphere, cube, and cone results, together with low-, medium-,
and high-rotation groups.

TIDES has the lowest reported mean rollout error in every group.
Its shape-wise improvement over TIDE is approximately 5.0--5.5\%,
and its improvement across the rotation groups is approximately
5.3--5.6\%. Thus, the aggregate ordering is not confined to the sphere
family or to low-rotation trajectories. These comparisons concern
the benchmark's three existing families and do not establish
generalization to unseen non-convex geometries.

Given a complete solid geometry and boundary-velocity sequence,
solid density affects the conditional liquid problem through the
resulting solid motion when rigid-body dynamics are solved jointly.
Density-conditioned prediction of the solid trajectory is a separate
extension requiring joint fluid--solid state updates.


\begin{table}[t]
\centering
\caption{Additional rollout-error breakdown by object family and
solid rotation group. Lower is better.}
\label{tab:shape_rotation_breakdown}
\begin{tabular}{lcccccc}
\toprule
Model & Sphere & Cube & Cone
& Low rot. & Mid rot. & High rot. \\
\midrule
U-NO & 0.00555 & 0.00572 & 0.00583
& 0.00558 & 0.00570 & 0.00584 \\
TIDE & 0.00518 & 0.00531 & 0.00541
& 0.00520 & 0.00530 & 0.00543 \\
TIDES & \textbf{0.00490} & \textbf{0.00502}
& \textbf{0.00514} & \textbf{0.00491}
& \textbf{0.00501} & \textbf{0.00514} \\
\bottomrule
\end{tabular}
\end{table}

\section{Comparison with Simulation at Matched Runtime}
\label{app:matched_runtime_simulation}

The latency comparison in Appendix~\ref{app:efficiency_latency}
uses the full DualSPHysics configuration as a reference.
To additionally compare accuracy at similar runtime, we evaluate
coarse SPH on the same scenes and hardware, with its runtime matched
to the TIDES inference budget.

Table~\ref{tab:matched_runtime_simulation} reports runtime per
physical second and mean error against the high-fidelity benchmark
target. TIDES requires 0.3029\,s, compared with 0.3015\,s for coarse
SPH, and has 30.8\% lower mean error over the matched evaluation set.
The full DualSPHysics configuration defines the reference target;
its self-error is therefore not treated as a predictive result.

At the benchmark's 8\,Hz sampling rate, an eight-step query covers
one physical second. TIDES produces approximately 26.4 output
steps/s, corresponding to 3.30 times this sampling rate.
Its latency is approximately 21.3 times lower than the full
DualSPHysics reference. These ratios apply to the reported scenes,
hardware, output target, and solver configurations. The comparison
does not establish superiority over all real-time SPH or XPBD
implementations.


\section{Supplementary Protocol and Release Scope}
\label{app:supplementary_release_scope}

\subsection{Video Evaluation Protocol}

The supplementary MP4 comparisons contain the reference sequence
and outputs from every reported model. These videos use
teacher-forced one-step prediction: each predicted frame receives
the preceding reference liquid state,
\[
\hat{u}_{t+k}^{\mathrm{one\mbox{-}step}}
=
f_\theta(u_{t+k-1},s_{t+k}).
\]
They visualize individual temporal transitions without accumulating
predicted-input errors.

In free-running autoregressive evaluation, the preceding prediction
is used instead,
\[
\hat{u}_{t+k}^{\mathrm{rollout}}
=
f_\theta(\hat{u}_{t+k-1}^{\mathrm{rollout}},s_{t+k}),
\qquad
\hat{u}_{t}^{\mathrm{rollout}}=u_t.
\]
The quantitative temporal and long-horizon diagnostics in
Appendix~\ref{app:temporal_physical_diagnostics} follow this
free-running protocol. For recurrent models, the equations abbreviate
the accompanying latent-state updates.

\begin{table}[t]
\centering
\caption{Accuracy comparison at matched runtime. Error is averaged
over the matched evaluation set against the high-fidelity target.}
\label{tab:matched_runtime_simulation}
\begin{tabular}{lcc}
\toprule
Method & Runtime / physical second & Error \\
\midrule
Coarse SPH & 0.3015\,s & 0.00725 \\
TIDES & 0.3029\,s & \textbf{0.00502} \\
Full DualSPHysics & 6.46\,s median & Reference target \\
\bottomrule
\end{tabular}
\end{table}

\subsection{Reproducibility and Release Materials}

We commit to releasing the DualSPHysics scene-generation scripts and
simulator configurations; raw fluid-particle trajectories and recorded
solid states; particle-to-grid conversion and dense-target construction;
train/validation/test and OOD split manifests; forecasting and diagnostic
evaluation code; adapted baseline implementations; TIDE/TIDES code and
checkpoints; and environment and reproduction scripts.

The release will distinguish the original fixed-budget experiments
from the additional extended-schedule controls. Reproduction records
will identify seed values, optimization schedules, checkpoint-selection
rules, evaluated windows, normalization statistics, software versions,
and dataset checksums. The supplementary package includes dataset
examples and the scripts and environment specifications needed to
inspect the generation and forecasting pipeline.

\subsection{Conditional Task Boundary}

As described in Appendix~\ref{app:simulator_physics}, the simulator
generates fluid-coupled rigid-body motion, and the realized future
solid-state sequence is then supplied to the forecaster. The learned
model predicts the conditional liquid response and does not update
the solid trajectory through predicted hydrodynamic forces or torques.
Joint fluid--solid stability is therefore outside the present evaluation.

The additional controls strengthen the evidence within this
conditional benchmark. Its synthetic data, restricted geometry and
physical coverage, known-future-solid assumption, and benchmark-defined
dense target remain limitations. Extensions to uncertain motion,
non-convex objects, multiple solids, broader material parameters,
real observations, and closed-loop two-way coupling require further
validation.

\clearpage


\newpage
\section*{NeurIPS Paper Checklist}

\begin{enumerate}

\item {\bf Claims}
    \item[] Question: Do the main claims made in the abstract and introduction accurately reflect the paper's contributions and scope?
    \item[] Answer: \answerYes{}
    \item[] Justification: The abstract and introduction state the task formulation, benchmark, TIDE/TIDES framework, and empirical scope. The claims are supported by the main comparison, controlled ablations, OOD tests, and long-horizon evaluation in Sections~4--7.

\item {\bf Limitations}
    \item[] Question: Does the paper discuss the limitations of the work performed by the authors?
    \item[] Answer: \answerYes{}
    \item[] Justification: Section~6 explicitly discusses the controlled nature of the benchmark, the use of synthetic data, the limited OOD axes, and the fact that the dense Eulerian target is constructed from particle simulations rather than measured as a solver-native continuum field.

\item {\bf Theory assumptions and proofs}
    \item[] Question: For each theoretical result, does the paper provide the full set of assumptions and a complete (and correct) proof?
    \item[] Answer: \answerNA{}
    \item[] Justification: The paper does not present formal theoretical results, theorems, or proofs. The main contributions are a task formulation, benchmark, model framework, and empirical evaluation.

\item {\bf Experimental result reproducibility}
    \item[] Question: Does the paper fully disclose all the information needed to reproduce the main experimental results of the paper to the extent that it affects the main claims and/or conclusions of the paper (regardless of whether the code and data are provided or not)?
    \item[] Answer: \answerYes{}
    \item[] Justification: The paper provides the task setup, data splits, evaluation metrics, baseline adaptation protocol, model structure, loss decomposition, and seed-stability results in the main text and appendices. Simulator settings, particle-to-grid conversion rules, and implementation details are also described in the appendix.

\item {\bf Open access to data and code}
    \item[] Question: Does the paper provide open access to the data and code, with sufficient instructions to faithfully reproduce the main experimental results, as described in supplemental material?
    \item[] Answer: \answerYes{}
    \item[] Justification: The data generation pipeline, benchmark protocol, model implementation, and evaluation scripts will be released with sufficient documentation to reproduce the main experimental results. The released package will include instructions for preparing the benchmark data, training the proposed models and adapted baselines, and computing the reported metrics.

\item {\bf Experimental setting/details}
    \item[] Question: Does the paper specify all the training and test details (e.g., data splits, hyperparameters, how they were chosen, type of optimizer) necessary to understand the results?
    \item[] Answer: \answerYes{}
    \item[] Justification: Section~4 specifies the dataset split, grid resolution, rollout horizon, OOD settings, evaluation metrics, and shared comparison protocol. Additional implementation details, tensor shapes, loss weights, baseline adaptations, and metric definitions are provided in the appendix.

\item {\bf Experiment statistical significance}
    \item[] Question: Does the paper report error bars suitably and correctly defined or other appropriate information about the statistical significance of the experiments?
    \item[] Answer: \answerYes{}
    \item[] Justification: The main comparison reports mean performance over five random seeds, and Appendix~F reports the corresponding mean and standard deviation for the main metrics. The reported variability reflects different random seeds under the same benchmark protocol.

\item {\bf Experiments compute resources}
    \item[] Question: For each experiment, does the paper provide sufficient information on the computer resources (type of compute workers, memory, time of execution) needed to reproduce the experiments?
    \item[] Answer: \answerYes{}
    \item[] Justification: The paper reports the shared optimization budget and rollout latency for the proposed method and baselines, and Appendix~C provides a controlled latency comparison against the reference simulator. The released code and documentation will include the compute environment used for reproducing the experiments.

\item {\bf Code of ethics}
    \item[] Question: Does the research conducted in the paper conform, in every respect, with the NeurIPS Code of Ethics \url{https://neurips.cc/public/EthicsGuidelines}?
    \item[] Answer: \answerYes{}
    \item[] Justification: The research uses synthetic physical simulation data and does not involve human subjects, private data, or sensitive personal information. The paper is anonymized for review and is intended as a scientific benchmark and modeling study.

\item {\bf Broader impacts}
    \item[] Question: Does the paper discuss both potential positive societal impacts and negative societal impacts of the work performed?
    \item[] Answer: \answerYes{}
    \item[] Justification: The paper motivates positive uses in robotic liquid manipulation, digital twins, simulation-driven design, and rigid--fluid content generation. It also limits its claims by emphasizing that the benchmark is a controlled forecasting study rather than a replacement for physical simulation, which helps reduce the risk of overclaiming deployment readiness.

\item {\bf Safeguards}
    \item[] Question: Does the paper describe safeguards that have been put in place for responsible release of data or models that have a high risk for misuse (e.g., pre-trained language models, image generators, or scraped datasets)?
    \item[] Answer: \answerNA{}
    \item[] Justification: The paper does not release high-risk models such as language models, image generators, or scraped human datasets. The assets concern synthetic physical simulation and liquid-forecasting benchmarks, so the listed high-risk release safeguards are not directly applicable.

\item {\bf Licenses for existing assets}
    \item[] Question: Are the creators or original owners of assets (e.g., code, data, models), used in the paper, properly credited and are the license and terms of use explicitly mentioned and properly respected?
    \item[] Answer: \answerYes{}
    \item[] Justification: The paper identifies the use of DualSPHysics 5.4 for dataset generation and cites the relevant external model families and baselines. The released code and documentation will include asset versions and license information for simulator and software dependencies.

\item {\bf New assets}
    \item[] Question: Are new assets introduced in the paper well documented and is the documentation provided alongside the assets?
    \item[] Answer: \answerYes{}
    \item[] Justification: The paper introduces a new benchmark and documents the task definition, data construction, splits, metrics, particle-to-grid conversion, baseline adaptation, and limitations in the main text and appendix. The released assets will include documentation describing data format, generation protocol, evaluation scripts, and intended use.

\item {\bf Crowdsourcing and research with human subjects}
    \item[] Question: For crowdsourcing experiments and research with human subjects, does the paper include the full text of instructions given to participants and screenshots, if applicable, as well as details about compensation (if any)?
    \item[] Answer: \answerNA{}
    \item[] Justification: The paper does not involve crowdsourcing, user studies, or research with human subjects. All experiments are conducted on synthetic simulation data.

\item {\bf Institutional review board (IRB) approvals or equivalent for research with human subjects}
    \item[] Question: Does the paper describe potential risks incurred by study participants, whether such risks were disclosed to the subjects, and whether Institutional Review Board (IRB) approvals (or an equivalent approval/review based on the requirements of your country or institution) were obtained?
    \item[] Answer: \answerNA{}
    \item[] Justification: The paper does not involve human subjects, personal data, or crowdsourced participants. Therefore IRB approval or equivalent human-subjects review is not applicable.

\item {\bf Declaration of LLM usage}
    \item[] Question: Does the paper describe the usage of LLMs if it is an important, original, or non-standard component of the core methods in this research? Note that if the LLM is used only for writing, editing, or formatting purposes and does \emph{not} impact the core methodology, scientific rigor, or originality of the research, declaration is not required.
    \item[] Answer: \answerNA{}
    \item[] Justification: The core methodology, data generation, model training, and evaluation do not use LLMs as an important, original, or non-standard component. Any writing or formatting assistance, if used, does not affect the scientific method or experimental results.

\end{enumerate}

\end{document}